\documentclass{article}
\usepackage[a4paper, margin=1in]{geometry}

\usepackage{microtype}
\usepackage{graphicx}
\usepackage[abs]{overpic}
\usepackage{subcaption}
\usepackage{booktabs}
\usepackage{diagbox}
\usepackage[affil-it]{authblk}

\usepackage{xcolor}
\definecolor{bahamablue}{RGB}{0, 104, 150}
\definecolor{apple}{RGB}{0, 104, 150}
\definecolor{vinrouge}{RGB}{153, 76, 102} 
\definecolor{citationblue}{RGB}{0, 102, 204}
\usepackage[
  colorlinks=true,       
  linkcolor=vinrouge,       
  citecolor=bahamablue,
  urlcolor=apple  
]{hyperref}
\newcommand{\refsubfigure}[1]{\textcolor{vinrouge}{#1}}

\usepackage{amsmath}
\usepackage{amssymb}
\usepackage{mathtools}
\usepackage{amsthm}
\theoremstyle{plain}

\theoremstyle{definition}

\theoremstyle{remark}

\usepackage{commath}
\usepackage{dsfont}
\usepackage{makecell}

\newcommand{\y}{\mathbf{y}}
\newcommand{\Y}{\mathbf{Y}}
\newcommand{\thetab}{\boldsymbol{\theta}}
\renewcommand{\c}{\mathbf{c}}
\newcommand{\C}{\mathbf{C}}
\renewcommand{\S}{\mathbf{S}}
\newcommand{\E}{\mathbb{E}}

\newcommand{\N}{\mathcal{N}}
\renewcommand{\d}{\,\mathrm{d}}
\usepackage{dirtytalk}

\usepackage{siunitx}
\usepackage[left]{lineno}

\title{Overcoming Selection Bias in Statistical Studies With Amortized Bayesian Inference}
\author[1,2]{Jonas Arruda}
\author[3,4]{Sophie Chervet}
\author[5,6]{Paula Staudt}
\author[7,8,9,10]{Andreas Wieser}
\author[7,8,9,11]{Michael Hoelscher}
\author[12,13,14]{Isabelle Sermet-Gaudelus}
\author[6,15]{Nadine Binder}
\author[3,4]{Lulla Opatowski}
\author[1,2]{Jan Hasenauer\thanks{Corresponding author: \href{mailto:jan.hasenauer@uni-bonn.de}{jan.hasenauer@uni-bonn.de}}}
\affil[1]{Bonn Center for Mathematical Life Sciences, University of Bonn, Bonn, Germany}
\affil[2]{Life \& Medical Sciences Institute, University of Bonn, Bonn, Germany}
\affil[3]{Epidemiology and Modeling of Antibiotic Evasion Unit, Institut Pasteur, Paris, France}
\affil[4]{Université de Versailles Saint-Quentin-en-Yvelines, Université Paris Saclay, Inserm U1018, Team Infectious Diseases, Interactions and Antimicrobial Resistance, Paris, France}
\affil[5]{Institute of Medical Biometry and Statistics, Faculty of Medicine and Medical Center, University of Freiburg, Freiburg, Germany}
\affil[6]{Freiburg Center for Data Analysis, Modeling and AI, University of Freiburg, Freiburg, Germany}
\affil[7]{Institute of Infectious Diseases and Tropical Medicine, LMU University Hospital, Munich, Germany}
\affil[8]{German Center for Infection Research, Partner Site Munich, Munich, Germany}
\affil[9]{Fraunhofer Institute ITMP, Immunology, Infection and Pandemic Research, Munich, Germany}
\affil[10]{Max von Pettenkofer Institute, LMU Munich, Munich, Germany}
\affil[11]{Unit Global Health, Helmholtz Zentrum München, German Research Center for Environmental Health (HMGU), Neuherberg, Germany}
\affil[12]{Centre de Référence Maladies Rares, Mucoviscidose et Maladies Apparentées, Site Constitutif Pédiatrique, Hôpital Necker Enfants Malades, Paris, France}
\affil[13]{Université de Paris, CNRS, INSERM, Institut Necker-Enfants Malades, Paris, France}
\affil[14]{European Rare Disease Network–Lung, Frankfurt, Germany}
\affil[15]{Institute of General Practice/Family Medicine, Faculty of Medicine and Medical Center, University of Freiburg, Freiburg, Germany}

\begin{document}

\maketitle

\begin{abstract}
Selection bias arises when the probability that an observation enters a dataset depends on variables related to the quantities of interest, leading to systematic distortions in estimation and uncertainty quantification. 
For example, in epidemiological or survey settings, individuals with certain outcomes may be more likely to be included, resulting in biased prevalence estimates with potentially substantial downstream impact.
Classical corrections, such as inverse-probability weighting or explicit likelihood-based models of the selection process, rely on tractable likelihoods, which limits their applicability in complex stochastic models with latent dynamics or high-dimensional structure.
Simulation-based inference enables Bayesian analysis without tractable likelihoods but typically assumes missingness at random and thus fails when selection depends on unobserved outcomes or covariates. 
Here, we develop a bias-aware simulation-based inference framework that explicitly incorporates selection into neural posterior estimation.
By embedding the selection mechanism directly into the generative simulator, the approach enables amortized Bayesian inference without requiring tractable likelihoods. 
This recasting of selection bias as part of the simulation process allows us to both obtain debiased estimates and explicitly test for the presence of bias. The framework integrates diagnostics to detect discrepancies between simulated and observed data and to assess posterior calibration.
The method recovers well-calibrated posterior distributions across three statistical applications with diverse selection mechanisms, including settings in which likelihood-based approaches yield biased estimates. These results recast the correction of selection bias as a simulation problem and establish simulation-based inference as a practical and testable strategy for parameter estimation under selection bias.
\end{abstract}

\section*{Introduction}

Population-level inference is central across many disciplines, including epidemiology \cite{rothman2008modern}, pharmacology \cite{strom2019pharmacoepidemiology}, medicine \cite{forrest2025using}, and economics \cite{heckman1979sample}.
However, in many such settings, the ability to generalize to the overall population cannot be guaranteed due to structural features of study design or data collection \cite{rudolph2023defining}. 
For instance, if women are less likely to participate in a seroprevalence study but are more likely to be infected, the sample will overrepresent men, and the estimated prevalence will be biased downward.
Such lack of representativeness is commonly referred to as selection bias and arises when the probability that an observation is included in a dataset depends on characteristics related to the quantities of interest, such as observed covariates, unmeasured confounders, outcome-dependent sampling, informative censoring, or latent variables that jointly influence inclusion in the study and outcomes \cite{cochran1977sampling, kleinbaum1981selection, rothman2008modern, smith2020selection}.
When selection bias is ignored, analyses can yield systematically biased inference results \cite{little2019statistical}. 
At the same time, multiple sources of bias are rarely modeled jointly, and selection bias is often neglected relative to other sources such as misclassification or uncontrolled confounding \cite{petersen2021systematic}.
With modern large-scale studies, where random errors diminish as sample size increases, systematic distortions arising from selection and study design can become the dominant source of estimation errors \cite{kaplan2014big}.

Most approaches to mitigating selection bias rely on explicitly modeling the data-generating and sampling processes under assumptions that yield tractable likelihoods or correction formulas.
Methods such as inverse probability weighting \cite{horvitz1952generalization}, the Heckman correction \cite{heckman1979sample}, propensity score techniques \cite{austin2011introduction}, and instrumental variables approaches \cite{StreeterLin2017} all require parametric structures or tractable representations of the selection mechanism to adjust estimates.
Bayesian formulations extend these paradigms by jointly modeling outcome and selection processes within a probabilistic framework \cite{copas1997inference, scharfstein2003incorporating, linero2018bayesian, kawabata2024accounting}. 
However, they likewise depend on tractable likelihoods for both the outcome and the selection process. 
These requirements impose restrictive parametric assumptions and limit applicability when the underlying processes involve latent stochastic dynamics, nonlinear interactions, or high-dimensional hidden states that preclude tractable computation. 
As statistical models become more complex, the necessity of tractable likelihoods becomes a central bottleneck, rendering bias correction infeasible.

Simulation-based inference (SBI) has emerged as a general framework for Bayesian inference in models with intractable likelihoods by approximating posterior distributions only using simulations generated by a statistical model. 
A broad class of approaches relies on neural density estimation to approximate the posterior distribution, including neural posterior estimation, neural likelihood estimation, and neural ratio estimation, and has been shown to be effective across various different applications \cite{LueckmannBoe2021}. 
These methods can be applied either in a dataset-specific manner or in an amortized setting. In amortized Bayesian inference, a generative neural network called a neural posterior estimator (NPE) is trained once on simulated data and can then be reused across arbitrarily many datasets at test time without retraining.
Because inference relies only on the ability to simulate from the model, SBI methods have been developed to work with complex, high-dimensional, and implicit generative processes \cite{arruda2025diffusionSBI}. 
Recent extensions have improved the robustness of SBI to practical challenges such as outliers \cite{SchaelteAla2021, bharti2026amortised, khoo2026minimum}, domain shifts \cite{elsemueller2025does}, and certain forms of missing data \cite{WangHas2024, gloeckler2024all}.

Despite these advances, SBI methods generally assume that the observed data constitute unbiased realizations of the modeled generative process. In practice, however, the data available for inference often result from structured selection mechanisms. 
When such mechanisms are not explicitly represented in the simulator, applying SBI to selectively observed data can lead to systematically biased posterior estimates and miscalibrated uncertainty. 
Detecting such violations is itself challenging, as the true parameters are typically unknown and standard posterior predictive checks and coverage tests cannot reveal biases induced by the selection mechanism.
To date, no amortized framework explicitly accounts for outcome-dependent or dynamically coupled selection mechanisms while simultaneously providing tests to detect bias and assess posterior calibration, leaving bias-aware likelihood-free inference under structured selection an open methodological challenge.

 To address this limitation, we develop a bias-aware amortized Bayesian inference framework that embeds the selection mechanism directly within the generative model. Selection, censoring, and missingness are formalized as structured observation operators acting on latent population-level processes, such that the population dynamics and the bias-inducing selection process are jointly simulated and NPEs are trained to infer population parameters from biased data. This likelihood-free construction requires only forward simulations and therefore avoids the need for tractable likelihoods. Because the framework can be amortized across similar selection mechanisms, it enables inference under heterogeneous and complex study designs within a unified model. Simulating both the data-generating and selection mechanism further allows systematic assessment of posterior calibration using simulation-based calibration (SBC) \cite{TaltsBet2020} and classifier two-sample tests (C2ST) \cite{lopez-paz2017revisiting}, which enable assessment of bias induced by a selection mechanism and detection of bias in real datasets. We evaluate the approach in three representative epidemiological settings reflecting distinct forms of structured selection: prevalence estimation under biased sampling with partially missing outcomes, incidence estimation under informative censoring due to death with coupled event and survival processes, and transmission modeling in which covariates and infection outcomes jointly influence study inclusion. Across these scenarios, the proposed approach yields calibrated inference even in regimes where classical corrections become unreliable due to model complexity or intractable likelihoods.

\section*{Results}

\begin{figure}[t]
    \centering
    \includegraphics[width=\linewidth]{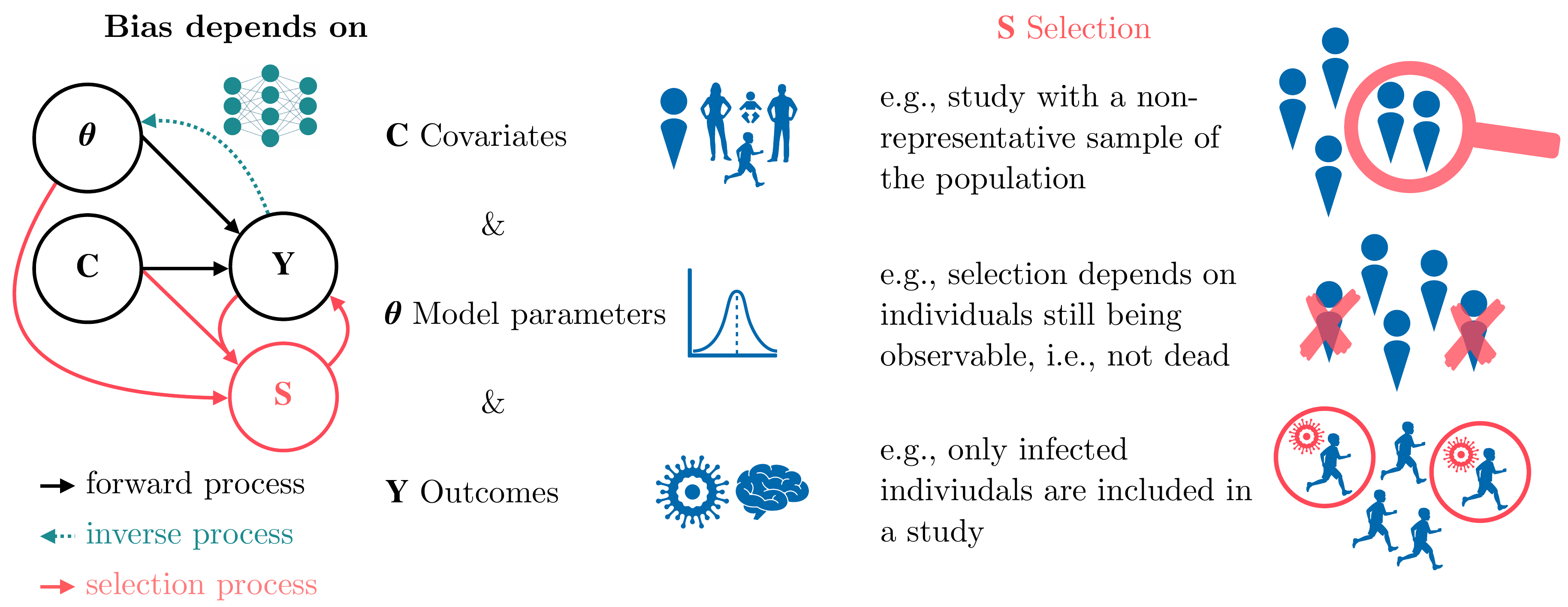}
    \caption{
    \emph{Different sources of selection bias in statistical studies.}
    The forward process generates outcomes from model parameters and covariates, with selection acting as an additional component that determines which individuals are observed. Selection may depend on covariates alone (e.g., a non-representative sample), on latent states such as survival and hence parameters (e.g., individuals must be alive to remain observable), or on outcomes (e.g., only infected individuals are enrolled). The inverse process must account for the selection in the forward process for bias-aware inference.
    }
    \label{fig:overview}
\end{figure}

\subsection*{A Framework for Bias-Aware Amortized Bayesian Inference Under Selection}

To provide bias-aware inference under selection, we propose an amortized Bayesian inference framework that explicitly incorporates the selection mechanism into the generative model. The framework builds on SBI using neural posterior estimation and addresses the challenge that observational datasets may arise from selection mechanisms that filter the underlying population. We therefore formulate the inferential target as the conditional distribution
\begin{equation*}
    p(\thetab \mid \Y, \S=1; \C),
\end{equation*}
where outcomes $\Y$ arise from a model with parameters $\thetab$ and covariates $\C$, and inclusion is governed by a binary selection process $\S$.
The process $\S$ is an outcome-, covariate-, or parameter-dependent stochastic selection process (\autoref{fig:overview}). In this setting, the observed dataset constitutes a filtered realization of the underlying generative process.

A key innovation of our approach is to embed the selection mechanism directly within the forward simulation used for inference. Parameters and covariates are sampled from the prior, outcomes are generated from a mechanistic population model, and the selection operator is subsequently applied to obtain the observed sample. Training the NPE on retained observations yields simulations from the joint distribution
$p(\thetab, \Y, \S=1, \C)$, so that the learned posterior approximation directly targets the correct conditional distribution under the assumed selection mechanism. Because this construction relies only on forward simulations, it avoids the need for tractable likelihoods for either the outcome model or the selection process (see \hyperref[sec:abi]{Methods}).

While the NPE is trained on simulated data, successful bias removal on real data needs to be validated. 
The amortized nature of the framework enables repeated fast inference and hence systematic validation of this aspect by repurposing established simulation-based diagnostics for the assessment of selection bias.
First, SBC is used to assess whether posterior ranks of ground-truth parameters are uniformly distributed across repeated simulations from the joint model, thereby verifying calibration with respect to the selection mechanism. 
Second, a C2ST is integrated to evaluate whether samples drawn from the learned posterior together with the observed data are statistically indistinguishable from parameter-data pairs generated directly from the joint distribution during training.
Together, these diagnostics formalize the criteria under which inference is considered calibrated and bias-aware.

The trained bias-aware NPE defines a global mapping from observed datasets to parameter distributions, and, due to inference being amortized, it can be evaluated across heterogeneous selection regimes without retraining. 
This enables systematic analysis of bias, calibration, and robustness under distinct forms of selection. In the following sections, we train bias-aware NPEs on three representative classes of selection mechanisms and examine the framework's inferential behavior in each setting.

\subsection*{Amortized Inference Reduces Selection Bias in an Incomplete Data Setting}

\begin{figure}[t]
\centering
\begin{subfigure}{0.49\linewidth}  
    \centering
    \begin{overpic}[width=\linewidth]{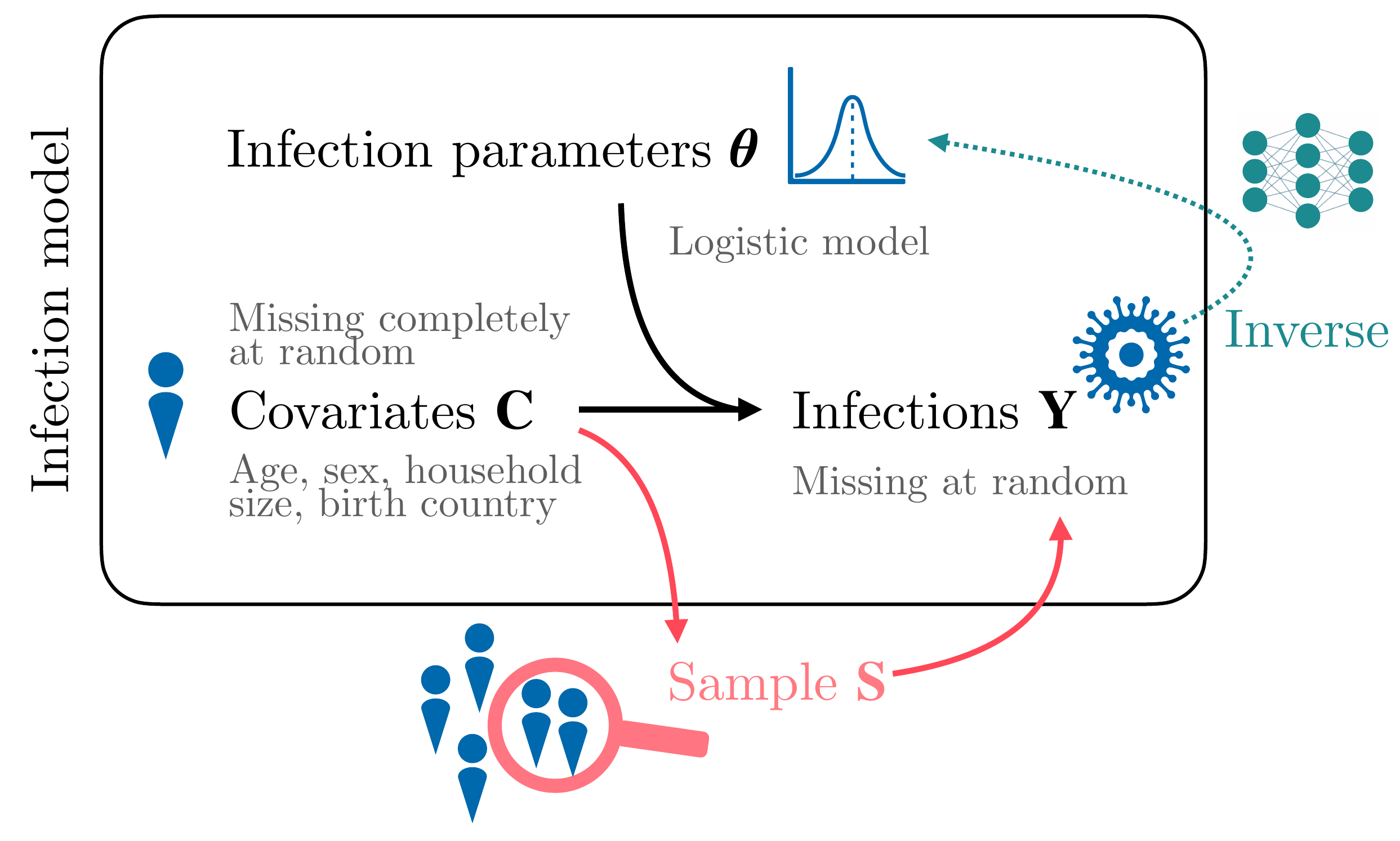}
        \put(-3,135){\textbf{A}}
    \end{overpic}
    \vspace{0.2em}
\end{subfigure}\hfill
\begin{subfigure}{0.49\linewidth}  
    \centering
    \begin{overpic}[width=\linewidth]{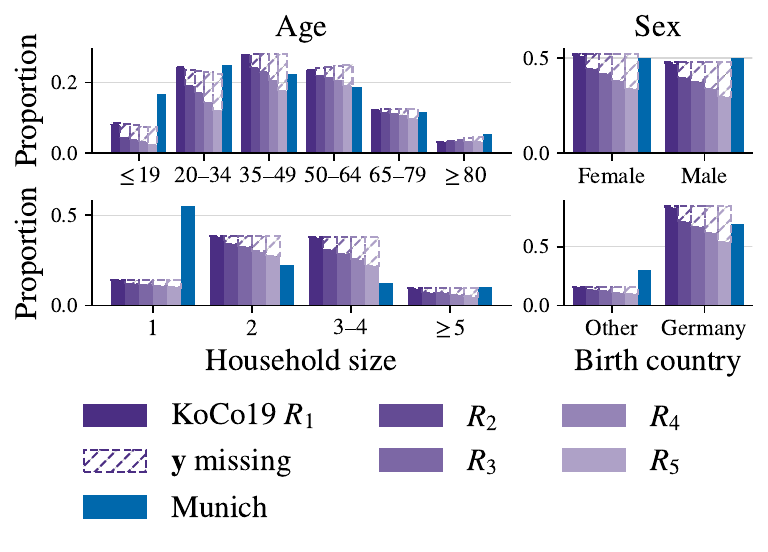}
        \put(-7,150){\textbf{B}}
    \end{overpic}
\end{subfigure}
\begin{subfigure}{0.49\linewidth}   
    \centering
    \begin{overpic}[width=\linewidth]{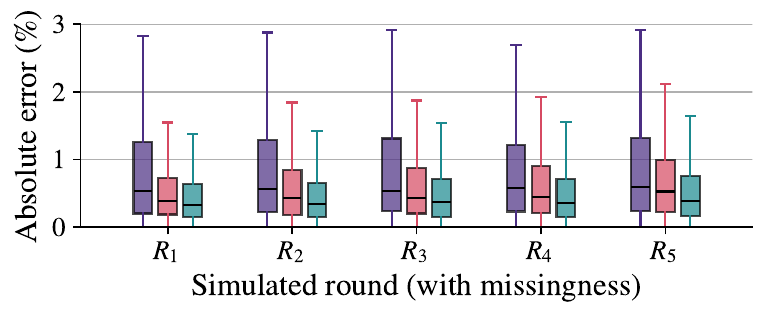}
        \put(-3,92){\textbf{C}}
    \end{overpic}
\end{subfigure}\hfill
\begin{subfigure}{0.49\linewidth}  
    \centering
    \begin{overpic}[width=\linewidth]{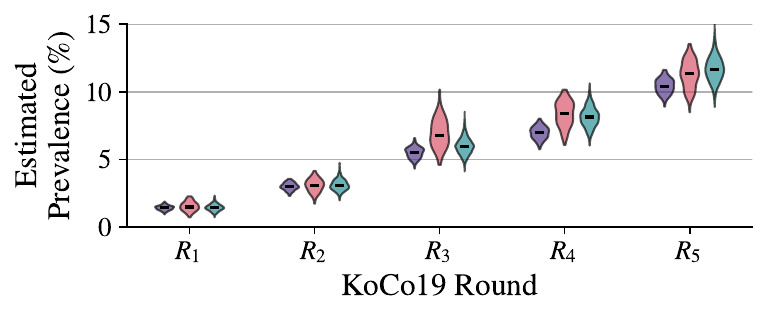}
        \put(-7,92){\textbf{D}}
    \end{overpic}
\end{subfigure}
\includegraphics[width=0.58\linewidth]{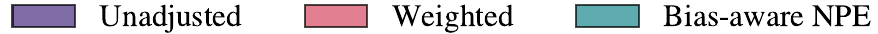}
\caption{
    \emph{Estimation of prevalence from biased population sample on simulated data and the KoCo19 Study.}
    (\textbf{A}) Visualization of the infection model and selection bias based on covariates.
    (\textbf{B}) Covariate distribution in the KoCo19 cohort for all 5 rounds compared to Munich, with indication for whom the infection status $\y$ is missing.
    (\textbf{C}) Absolute error of estimated prevalence across 1000 simulated datasets (including selection and missingness), each with five rounds of the KoCo19 Study using unadjusted counts of infections, inverse-probability weighting, and the mode of the posterior estimated with bias-aware NPE.
    (\textbf{D}) Estimated prevalence across five rounds of the KoCo19 Study using unadjusted counts of infections, inverse probability weighting, and bias-aware NPE.
    For the unadjusted and weighted approach, we bootstrapped the estimate, and for NPE, we show the full posterior.
    }
\label{fig:prevalence_results}
\end{figure}

To assess the proposed framework, we first consider disease prevalence estimation under sampling bias and covariate and outcome missingness (\autoref{fig:prevalence_results}\refsubfigure{A}). For this class of problems, several statistical methods are available for evaluating bias correction. Here, we consider data from the KoCo19 Study, which was conducted to estimate the seroprevalence of SARS-CoV-2 antibodies in the general population of Munich over time~\cite{radon2020protocol}. The study consists of 5 rounds with increasing missingness of outcomes (\autoref{fig:prevalence_results}\refsubfigure{B}).

We first evaluated the proposed framework in a controlled yet realistic simulation setting and subsequently applied it to the observed data. 
Ground-truth prevalence was obtained from a synthetic population matching Munich’s demographic structure, with infection outcomes simulated from a logistic model (see \hyperref[sec:prevalence]{Methods}). 
Non-representative samples were then generated by resampling according to the original study sample weights and by imposing the observed round-specific missingness patterns (\autoref{fig:prevalence_results}\refsubfigure{B}).
To assess the framework's performance, we trained a NPE on such biased subsamples paired with their corresponding true prevalence and estimated prevalence with an unadjusted estimator and inverse probability weighting (see \hyperref[sec:prevalence]{Methods}).
This bias-aware NPE learned to correct both for non-representative sampling and outcome missingness (\autoref{fig:prevalence_results}\refsubfigure{C}). 
On 1000 simulated datasets, bias-aware NPE yielded more accurate prevalence estimates than both the unadjusted estimator and inverse probability weighting, while retaining similarly fast inference due to amortization.
The latter two approaches relied here on complete case analysis and therefore exhibited bias and increased variance in the presence of missing outcomes and a reduced number of samples in the data (\autoref{fig:prevalence_results}\refsubfigure{B}).
When outcome missingness was removed from the simulation, the NPE recovered prevalence estimates that were indistinguishable from those obtained via inverse probability weighting, confirming that additional gains arise from explicitly accounting for missingness (\autoref{fig:prevalence_appendix}\refsubfigure{A}).

The application of the bias-aware NPE, the unadjusted estimator, and inverse probability weighting to the real study data revealed that for the first 2 rounds, the prevalence estimates were broadly consistent across the methods (\autoref{fig:prevalence_results}\refsubfigure{D}).
Here, the same amortized NPE trained exclusively on synthetic data was applied to all 5 study rounds without retraining.
In contrast, for rounds 3 and 5, the bias-aware NPE produced estimates that differed systematically from inverse probability weighting, as reflected by the different shapes of the uncertainty distributions. 
In round 3, inverse probability weighting yielded higher median prevalence estimates ($6.70\%$ vs.\ $5.96\%$), whereas in round 5 it produced lower median estimates compared to the bias-aware NPE ($11.19\%$ vs.\ $11.73\%$).
This discrepancy arises from the increasing degree of outcome missingness across rounds and suggests that inverse probability weighting alone is insufficient when missingness is informative (\autoref{fig:prevalence_results}\refsubfigure{C}).

The assessment of the reliability of the posterior using C2ST revealed that posterior samples generated by the bias-aware NPE were statistically indistinguishable from samples drawn from the joint simulator.
Moreover, on simulated validation datasets, the classifier operated at chance level, which confirms convergence of the neural network and calibration with respect to the simulated selection mechanism.
When applied to real study rounds, the classifier likewise performed at chance level (\autoref{fig:prevalence_appendix}\refsubfigure{B}), indicating that posterior samples were consistent with the assumed generative model and selection mechanism. These results support that the inferred posteriors remain calibrated despite the presence of sampling bias and outcome missingness.

\subsection*{Amortized Inference Enables Calibrated and Testable Time-to-Event Analysis Under Missing Disease Information}

\begin{figure}[t]
\centering
\begin{minipage}[t]{0.48\linewidth}
    \centering
    \vfill

    \begin{overpic}[width=\linewidth]{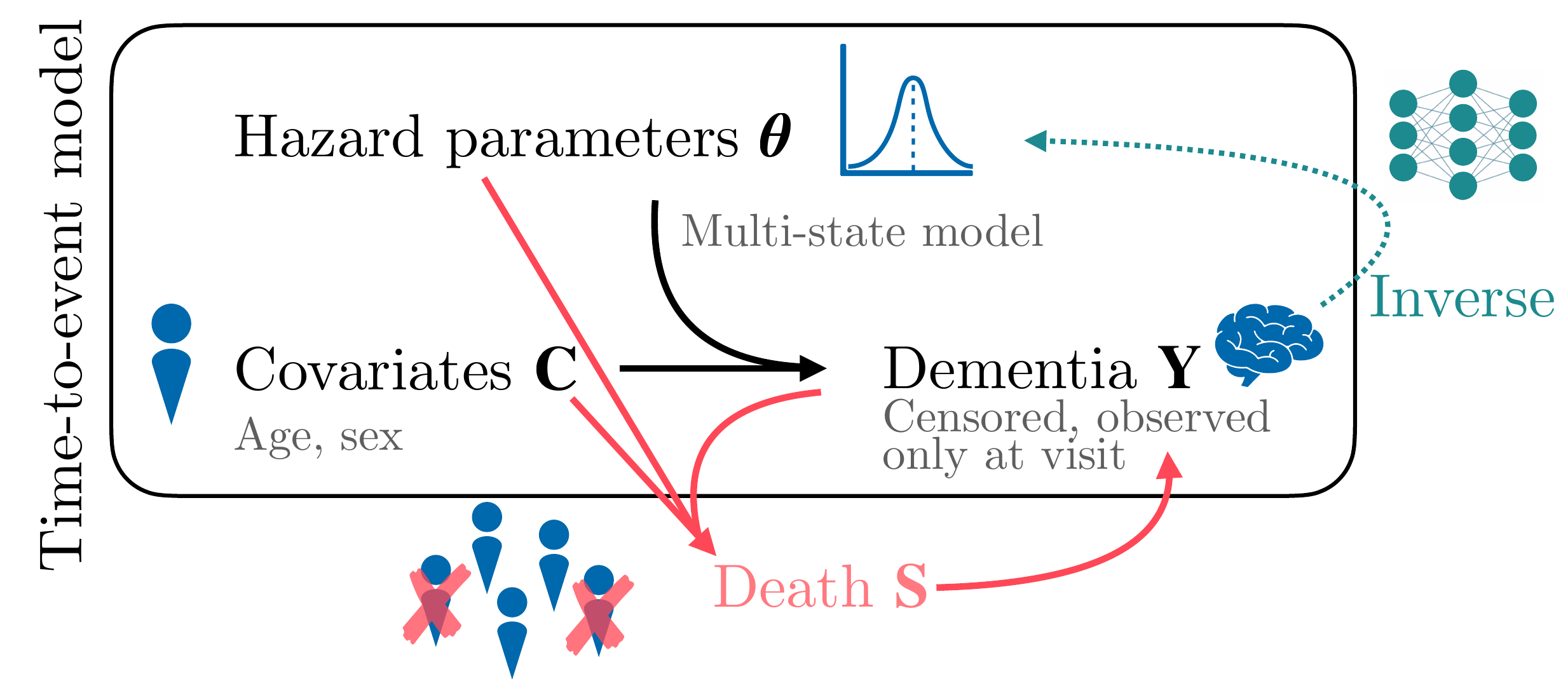}
        \put(-5,96){\textbf{A}}
    \end{overpic}

    \vspace{0.7em}

    \begin{overpic}[width=0.98\linewidth]{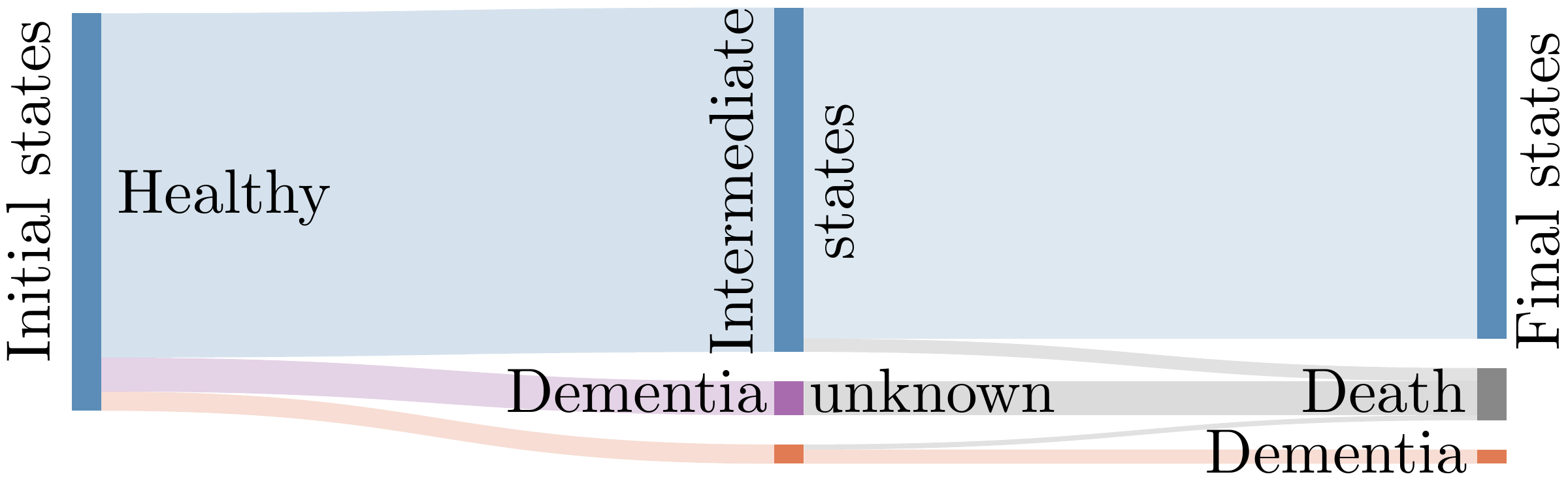}
        \put(-7,64){\textbf{B}}
    \end{overpic}

    \begin{overpic}[width=\linewidth]{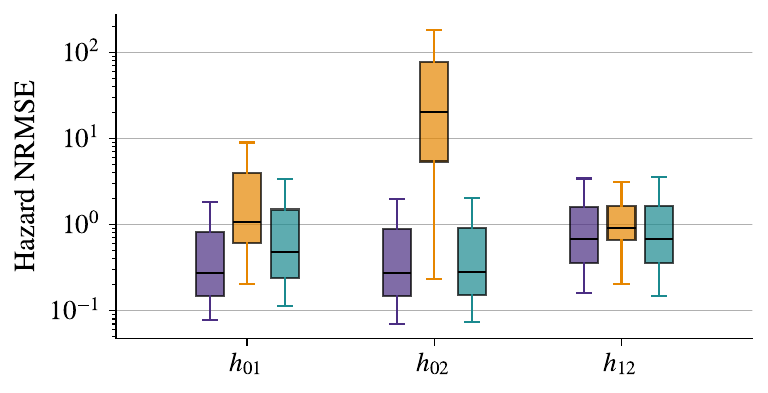}
        \put(-5,105){\textbf{C}}
    \end{overpic}
\end{minipage}%
\,
\begin{minipage}[t]{0.48\linewidth}
    \vfill
    \centering
    \begin{overpic}[width=\linewidth]{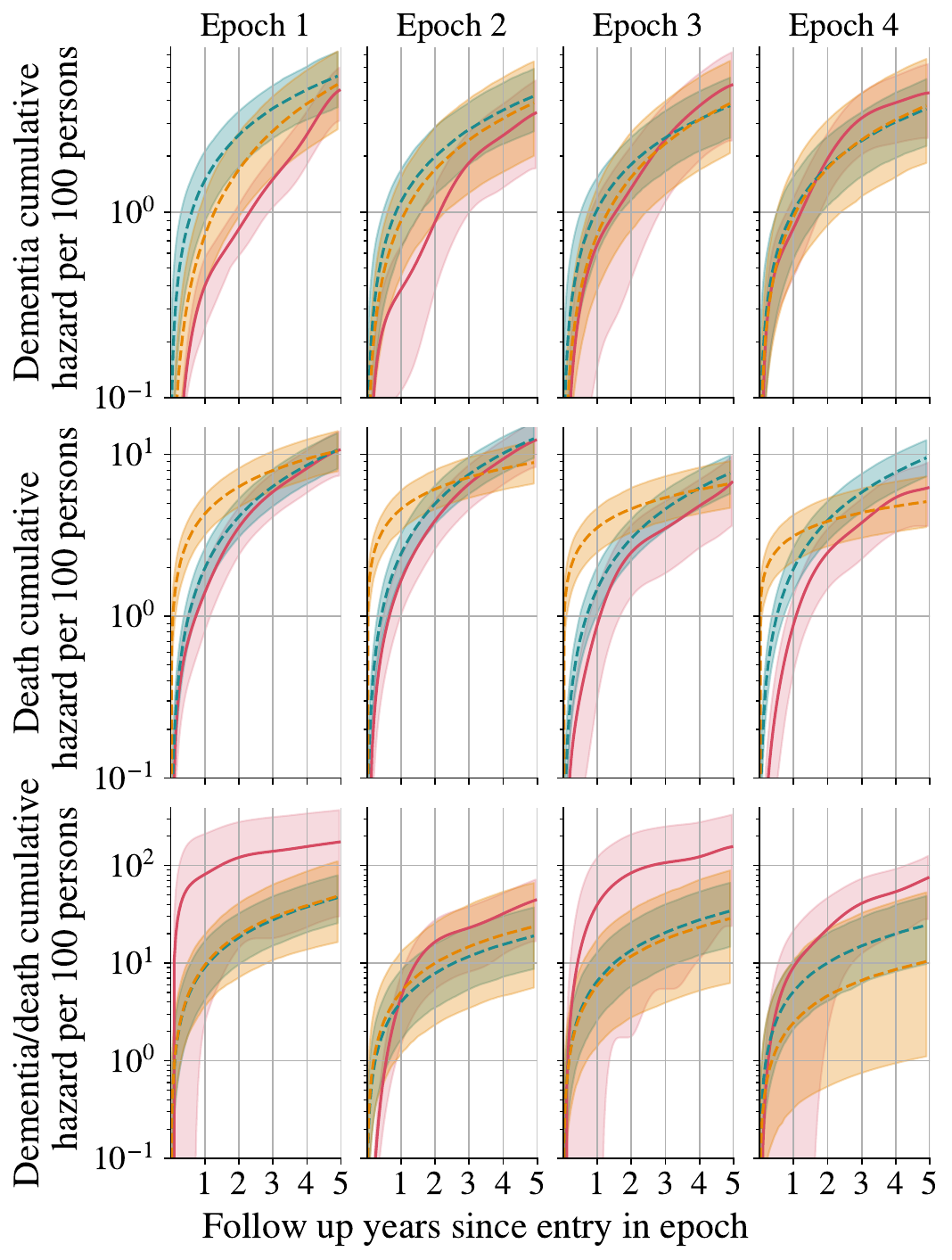}
        \put(0,289){\textbf{D}}
    \end{overpic}
\end{minipage}
\begin{subfigure}{0.70\linewidth}
    \centering
    \includegraphics[width=\linewidth]{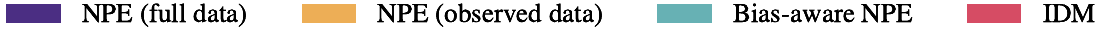}
\end{subfigure}

\caption{
\emph{Correcting for bias due to missing disease information because of death on simulations and the Framingham Heart Study.}
(\textbf{A}) Visualization of the time-to-event model and bias due to missing dementia because of death.
(\textbf{B}) Observable and unobservable transitions between states from over 2200 patients in the Framingham Heart Study summed over all 4 epochs. Parts of the transitions to death are interval-censored since last visit with unknown dementia status.
(\textbf{C}) Normalized root mean squared error (NRMSE) of the recovered transition hazard ($h_{kl}$: $0\to1$ dementia onset, $0\to2$ death without prior dementia, and $1\to2$ death after dementia) adjusted by covariates on simulated data with and without censoring. For each posterior sample the mean over time is computed and the error normalized by the mean of the corresponding ground truth hazard. Posterior samples are aggregated with the median, and the recovery error across datasets is shown.
(\textbf{D}) Cumulative hazards adjusted by covariates estimated using real data. We compared our bias-aware NPE approach (median cumulative hazard and $95\%$ credible intervals) against the NPE trained on full data and a penalized likelihood approach. 
For the latter, the maximum likelihood estimate is shown together with intervals obtained from the lower and upper intensity curves (interpreted as $95\%$ bounds of the hazard function) provided by the underlying IDM.
}
\label{fig:visit_censoring_results}
\end{figure}

\begin{figure}[t]
\centering
\begin{minipage}[b]{0.24\linewidth}
    \centering
    \begin{overpic}[width=\linewidth]{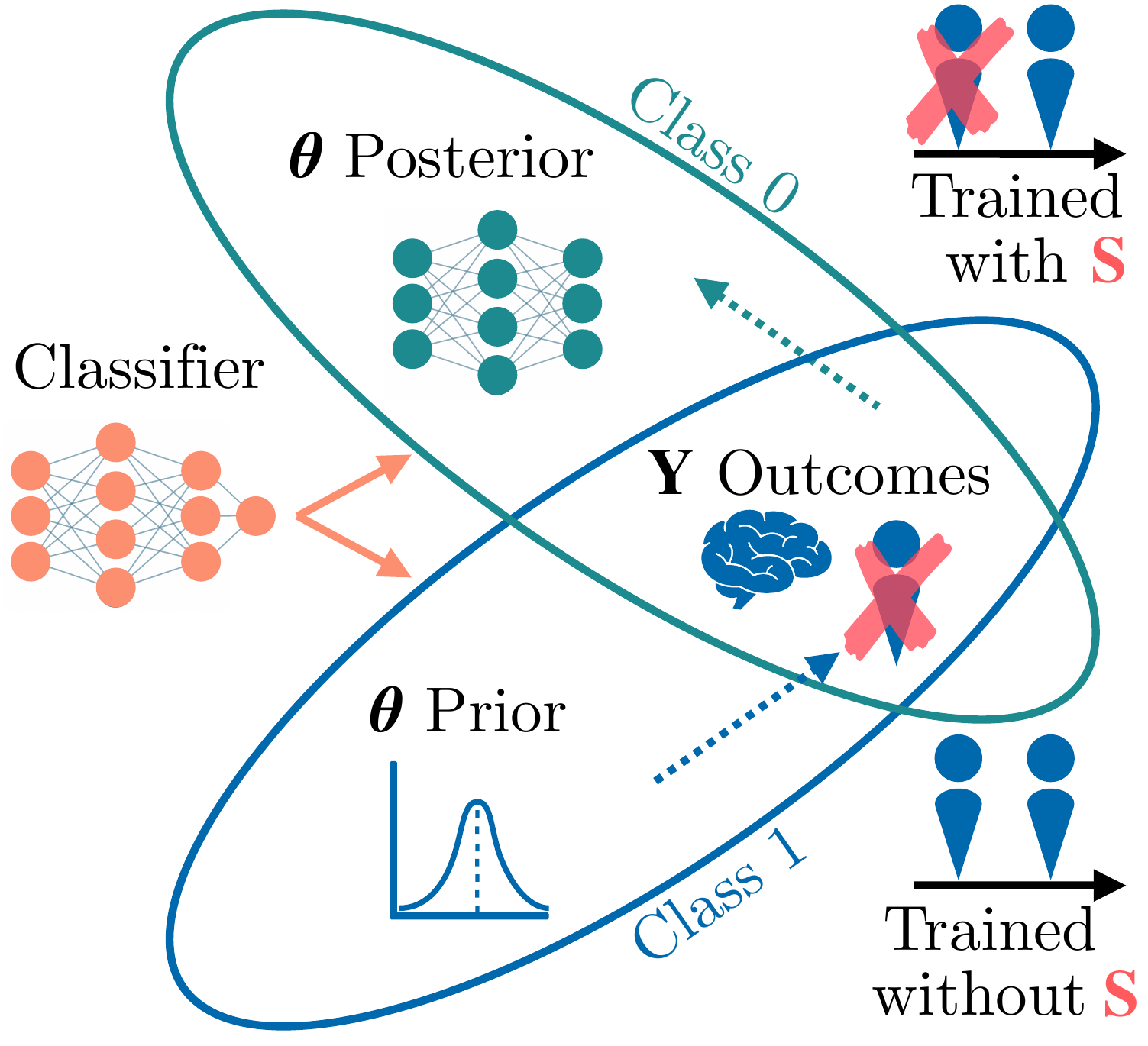}
        \put(0,100){\textbf{A}}
    \end{overpic}
    \vspace{1.2em}
\end{minipage}
\hfill
\begin{minipage}[b]{0.65\linewidth}
\centering
    \begin{overpic}[width=\linewidth]{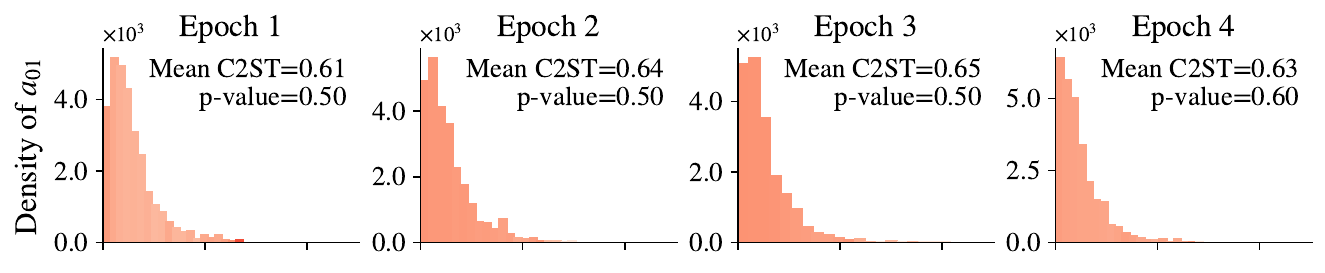}
    \put(0,53){\textbf{B}}
    \end{overpic}
    \begin{overpic}[width=\linewidth]{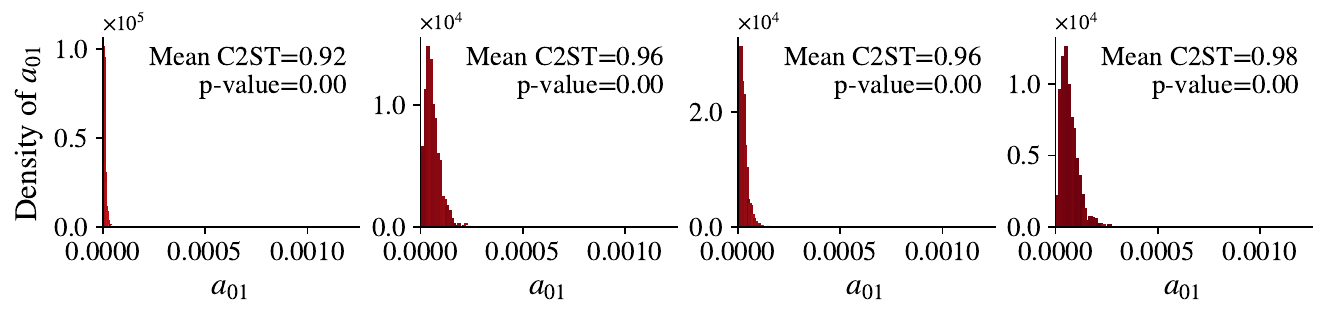}
    \put(0,65){\textbf{C}}
    \end{overpic}
\end{minipage}
\begin{minipage}[b]{0.08\linewidth}
    \centering
    \includegraphics[width=\linewidth]{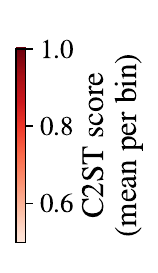}
    \vspace{3em}
\end{minipage}
\caption{
    \emph{Validating bias correction on the Framingham Heart Study.}
    (\textbf{A}) A classifier is trained on parameter-data pairs to predict if the pair belongs to the posterior or the joint.
    (\textbf{B}) Marginal posteriors of the bias-aware NPE for the baseline transition scale parameter $a_{01}$ (healthy to dementia) for all 4 epochs. Posterior samples were tested using a classifier-based diagnostic, where a C2ST score of 0.5 means that samples cannot be distinguished from samples of the joint distribution.
    (\textbf{C}) Marginal posteriors of a NPE trained only on full data. Colors indicate classifier-based diagnostics, also trained only on data that was not censored.
}
\label{fig:visit_censoring_results_c2st}
\end{figure}

To further assess the proposed framework, we consider time-to-event analysis under missing disease information due to death. In this setting, the event of interest cannot be observed once death occurs, which can introduce substantial bias in incidence estimates \cite{leffondre_interval-censored_2013,binder_missing_2014}. 
Such mechanisms can be seen as selection of observed data and can arise in any competing risk model and are particularly relevant for neurodegenerative diseases, such as dementia, where mortality competes with disease onset. 
We therefore consider data from the long-term population-based Framingham Heart Study, which has been used to study temporal trends in dementia incidence \cite{satizabal2016incidence}. We base the analysis on an established illness-death model (IDM) with states 
corresponding to healthy, dementia, and death, where death may obscure the transition from healthy to dementia (\autoref{fig:visit_censoring_results}\refsubfigure{A,B}).
We first evaluated the proposed framework in a controlled simulation setting and subsequently applied it to the observed data.

For the simulation study, we generated data by simulating trajectories from the IDM using covariates observed in the Framingham cohort (full data) and subsequently induced missing dementia information due to death (observed data) (see \hyperref[sec:cens_visit]{Methods}). 
The simulated data were analyzed using two different approaches: (1) a standard NPE trained on the full data and (2) the proposed bias-aware NPE trained on the data as observed. We analyzed recovery accuracy using the normalized root mean squared error of the covariate-adjusted transition hazards.

The assessment of the two methods using simulated data revealed clear performance differences.
For simulated data with induced missing dementia information, only the proposed bias-aware NPE achieved good performance (\autoref{fig:visit_censoring_results}\refsubfigure{C}).
Indeed, the bias-aware NPE recovered all transition hazards with accuracy comparable to the standard NPE trained on full data.
Consistent with this result, both models achieved similar C2ST statistics on their respective validation sets when the training regime matched the data-generating process (0.56 for observed data and 0.57 for full data). 
In contrast, applying the standard NPE to observed datasets substantially degraded hazard recovery. The largest errors occurred for the healthy-to-death $h_{02}$ transition, but bias was also evident in the remaining transitions (\autoref{fig:visit_censoring_results}\refsubfigure{C}). 
The dementia-to-death transition $h_{12}$ is in general hard to recover, as indicated by a low contraction of the hazard scale parameter (\autoref{tab:appendix_cens_visit_contraction}) due to the low number of observed transitions.
These results show that explicitly incorporating the observation process into the simulator is necessary for accurate parameter recovery under death-induced selection.

The real data were analyzed using a further approach: a tailored spline-based full likelihood approach derived from the same multi-state formulation, which was previously used to analyze the data \cite{binder2019multi}.
Applying all methods to the real Framingham data yielded estimates for the age- and sex-adjusted cumulative hazards for dementia, direct death, and death after dementia (\autoref{fig:visit_censoring_results}\refsubfigure{D}) as well as the impact of covariates (\autoref{fig:appendix_cens_vist_recovery}).
For the age- and sex-adjusted cumulative death hazards, we observed good agreement between the spline-based likelihood approach and our proposed bias-aware NPE, but not with the NPE trained on full data, in line with our previous findings on simulations.
In the first epoch, the cumulative dementia hazard estimated with the bias-aware NPE was slightly higher than with NPE trained on full data or the likelihood approach. For this transition, the NPE trained on full data showed more uncertainty than the other approaches.
All approaches showed high uncertainty for cumulative dementia-to-death transition hazards (\autoref{fig:visit_censoring_results}\refsubfigure{D}) due to the low number of known dementia-to-death transitions, reflecting limited identifiability of parameters governing death rates after dementia onset. 
For the impact of covariates, we found overall good agreement between methods with high uncertainty for the estimates related to sex of the NPE trained on full data~(\autoref{fig:appendix_cens_vist_recovery}).

We additionally analyzed the data using a naive Cox proportional hazards model, which censors individuals who die without a recorded dementia diagnosis at their last dementia-free visit. This ignores the possibility of dementia onset between the last dementia-free observation and death, leading to systematic bias \cite{binder2019multi}.
However, the resulting bias differs from that of the NPE trained on full data (\autoref{fig:appendix_cens_vist_recovery}). The Cox model tends to overestimate cumulative death hazards while underestimating dementia hazards, especially in later epochs, as deaths are fully observed but dementia events are partially unobserved. 
In contrast, the NPE trained on full data underestimates death hazards compared to all other methods due to a mismatch between training and observed data.
This shows that simulation-based methods can fail in unpredictable ways when simulated data and real data do not match.

To provide an additional verification of the reliability of the estimates obtained using the bias-aware NPE, we assessed the calibration of the inferred baseline transition scale parameter from healthy to dementia, $a_{01}$, across study epochs using C2STs (\autoref{fig:visit_censoring_results_c2st}\refsubfigure{A}). Posterior samples from the bias-aware NPE generated from the real data were not statistically distinguishable from samples drawn from the joint simulator, with C2ST scores around 0.65 across epochs (\autoref{fig:visit_censoring_results_c2st}\refsubfigure{B}). In contrast, posterior samples from the NPE trained exclusively on full data were clearly distinguishable from the joint distribution when evaluated on the real censored data, with C2ST scores close to 1.0 and statistically significant discrepancies (\autoref{fig:visit_censoring_results_c2st}\refsubfigure{C}). This mismatch indicates that ignoring censoring during training leads to an incorrect joint distribution and invalid posterior inference, while C2ST successfully detects the resulting bias even though neither the inference network nor the classifier was trained on censored data.

\subsection*{Enabling Unbiased Inference in Complex Stochastic Models and Study Designs with Intractable Likelihoods}

\begin{figure}[t]
\centering
\begin{subfigure}{0.39\linewidth}   
    \centering
    \begin{overpic}[width=\linewidth]{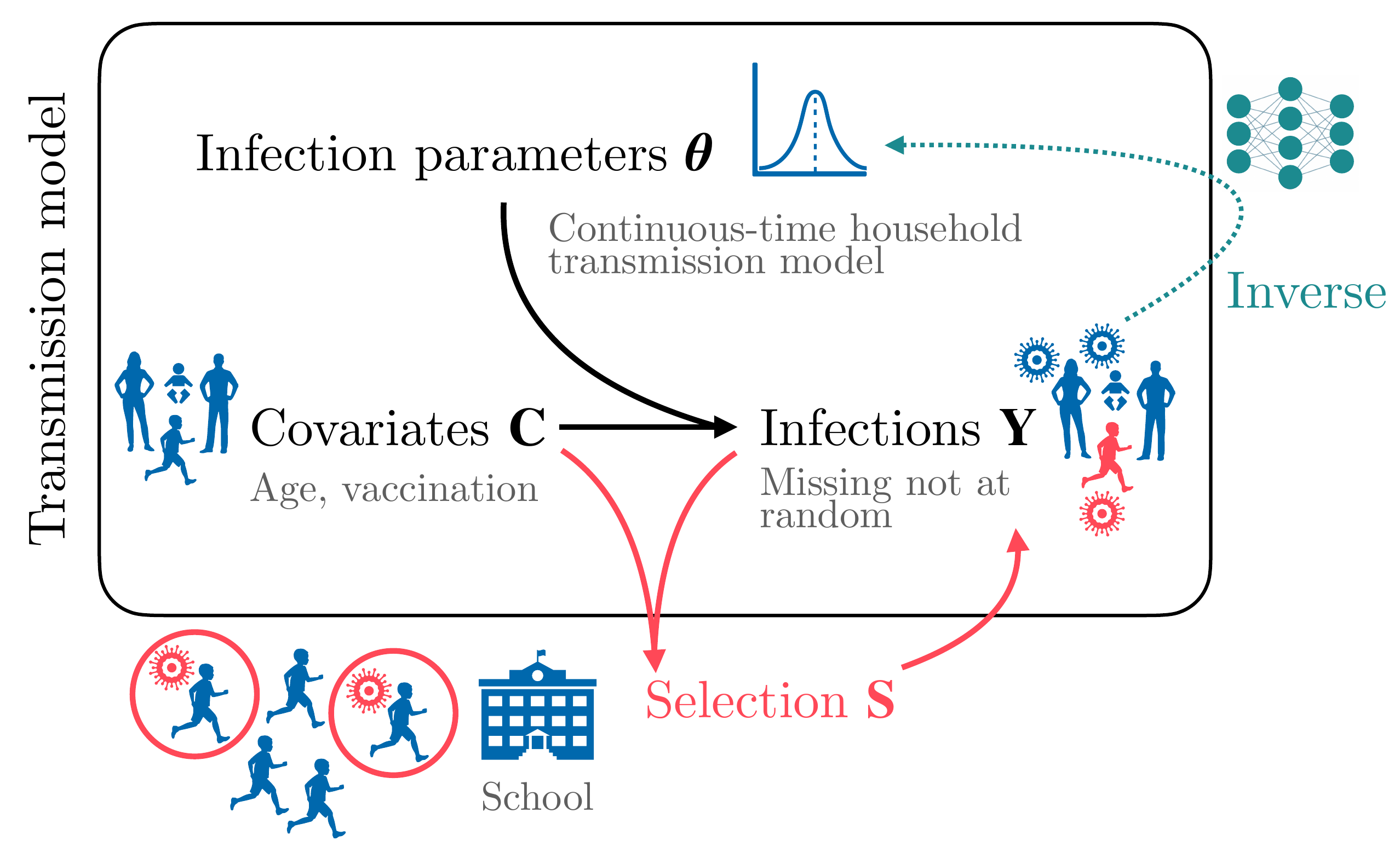}
    \put(0,116){\textbf{A}}
    \end{overpic}
    \vspace{0.2em}
\end{subfigure}\hfill
\begin{subfigure}{0.60\linewidth}  
    \centering
    \begin{overpic}[width=\linewidth]{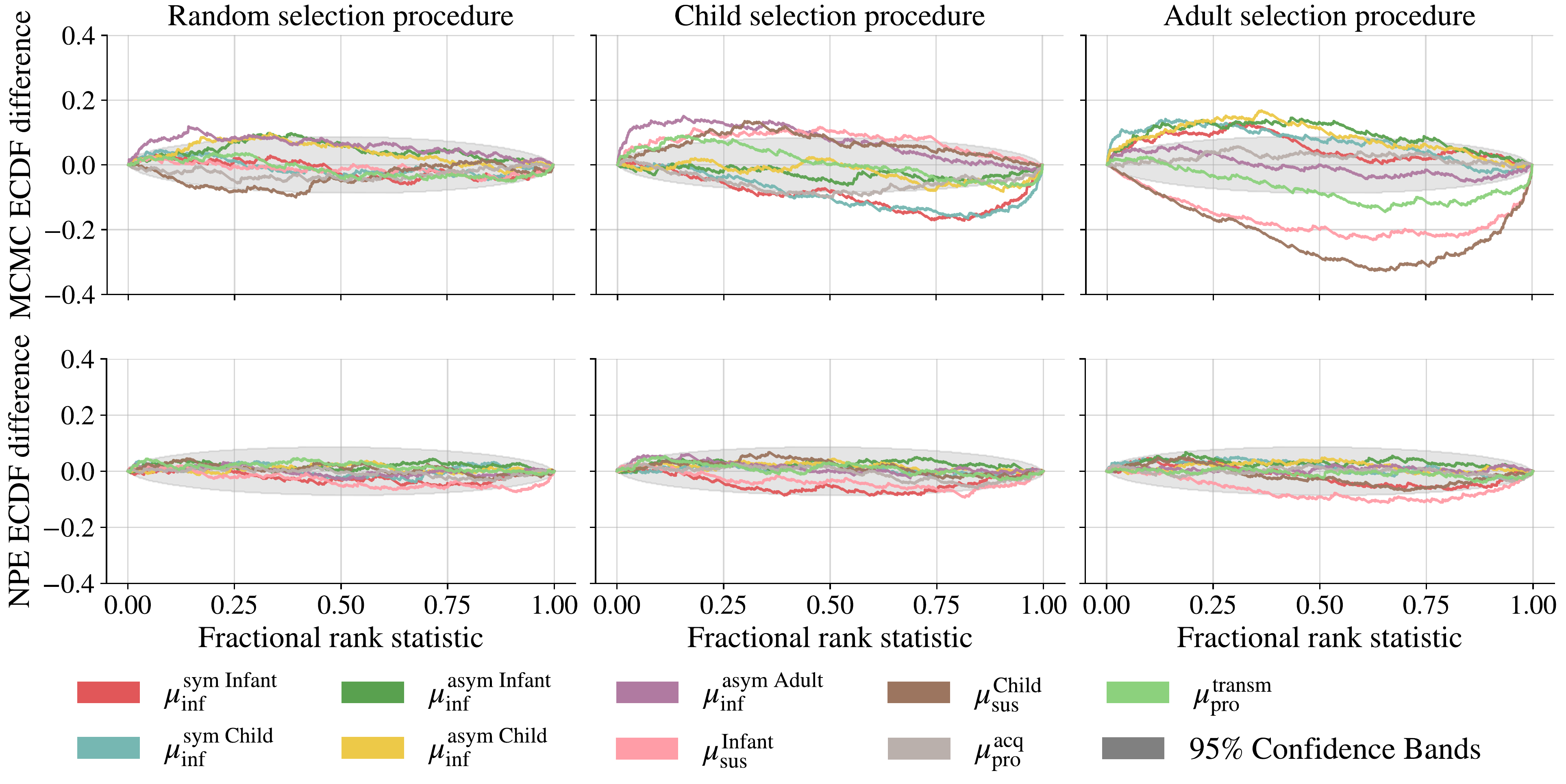}
    \put(-10,130){\textbf{C}}
    \end{overpic}
\end{subfigure}
\begin{subfigure}{0.35\linewidth}  
    \centering
    \begin{overpic}[width=\linewidth]{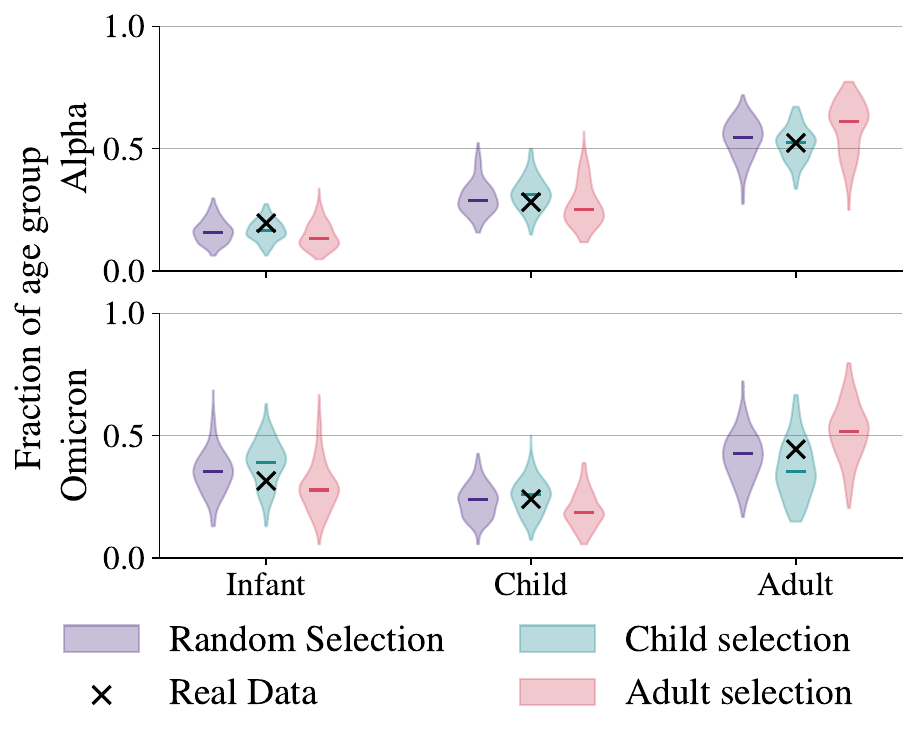}
    \put(0,122){\textbf{B}}
    \end{overpic}
\end{subfigure}\hfill
\begin{subfigure}{0.60\linewidth}   
    \centering
    \begin{overpic}[width=\linewidth]{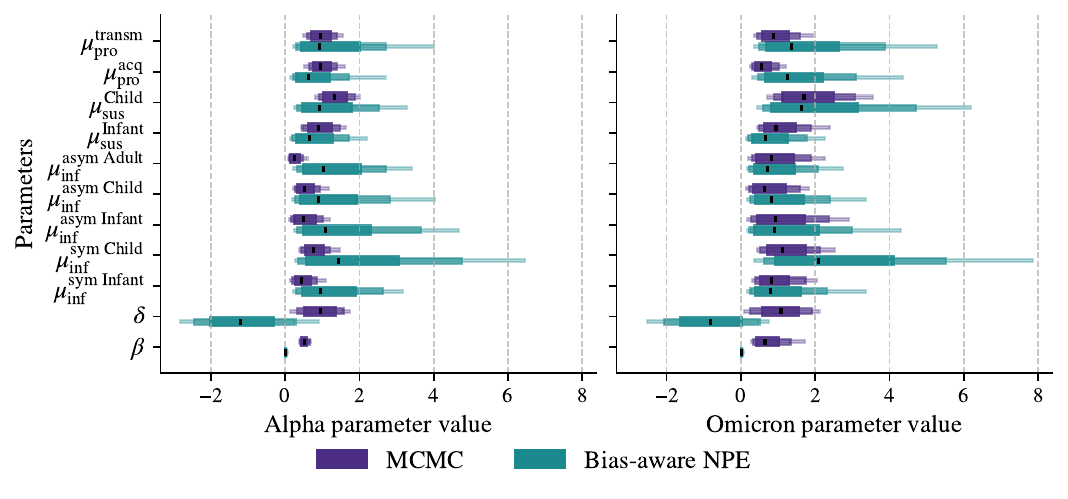}
    \put(-10,122){\textbf{D}}
    \end{overpic}
\end{subfigure}
\caption{
    \emph{Correcting for bias under multiple selection procedures on simulated data and the PedCovid Study.}
    (\textbf{A}) Visualization of selection bias based on outcome and covariates.
    (\textbf{B}) Fraction of household members who were the first to test positive in each age group, shown for real data and for different simulated selection mechanisms (infections occurring on the same day are assigned to the younger age group). A total of 300 datasets with identical parameters were simulated for each mechanism.
    (\textbf{C}) ECDF of MCMC and NPE plots for the 3 different selection procedures.
    (\textbf{D}) MCMC and NPE results for real data with child selection procedure for Alpha (left) and Omicron (right) variants.
}
\label{fig:pedcov_results}
\end{figure}

Having demonstrated reliable performance in applications with established estimators, we next assess whether the proposed framework can address problems for which a statistically coherent treatment remains challenging. We therefore consider inference under selection bias in complex stochastic simulation models where structured study inclusion distorts the observed data and the complexity of the underlying processes makes explicit likelihood-based correction infeasible (\autoref{fig:pedcov_results}\refsubfigure{A}).
As a representative example, we consider the PedCovid Study, a prospective longitudinal household study of SARS-CoV-2 transmission in which households were enrolled if there was a child that tested positive prior to the study inclusion date \cite{delaunaymoisan2022}. This outcome- and covariate-dependent inclusion criterion induces strong selection bias when inferring transmission parameters. 
We evaluate the proposed framework using both simulated datasets and the observed PedCovid data.

For the simulation study, household-level transmission data were generated using a mechanistic transmission model (see \hyperref[sec:pedcov]{Methods}). Observed households were replicated multiple times to construct synthetic populations, and infection dynamics were simulated conditional on parameters drawn from a prior distribution. To investigate the effect of study inclusion on parameter inference, datasets were generated under multiple selection schemes, including random household inclusion, child-dependent inclusion, and adult-dependent inclusion. Under random inclusion, households were eligible if the selected member tested positive before the inclusion date. Under child-dependent inclusion, the selected member additionally had to be younger than 18 years, whereas under adult-dependent inclusion, the selected member had to be 18 years or older. These inclusion criteria alter the composition of the observed households and thereby affect the distribution of infection characteristics, such as the age of the first infected household member (\autoref{fig:pedcov_results}\refsubfigure{B}), a key determinant of transmission dynamics given known differences in immune responses across age groups~\cite{manfroi2024preschool}.

Inference was performed using a bias-aware NPE trained on pairs of biased datasets and ground-truth parameters, with an explicit indicator of the selection mechanism included as part of the training data. For comparison, we additionally performed likelihood-based inference using Markov chain Monte Carlo (MCMC) with the No-U-Turn Sampler implemented in \texttt{Stan}~\cite{carpenter2017stan}. This likelihood-based formulation assumes random household inclusion, which is computationally tractable but does not account for outcome-dependent selection.
MCMC inference required over 12 hours of wall-clock time using 96 CPUs in parallel, whereas inference with the trained bias-aware NPE was performed in a matter of seconds on a single GPU.

Evaluation on simulated data showed that under random household inclusion, both MCMC and the bias-aware NPE recovered posterior distributions consistent with the ground truth. Under outcome- or covariate-dependent inclusion, the bias-aware NPE consistently recovered the true parameter values (\autoref{fig:appendix_pedcvo_recovery}\refsubfigure{A}).
However, MCMC produced systematically biased posterior estimates as reflected in the simulation-based calibration results (\autoref{fig:pedcov_results}\refsubfigure{C}): the empirical cumulative distribution functions (ECDFs) for MCMC deviated markedly from the uniform distribution under biased sampling, while those obtained from the amortized approach remained well calibrated. By conditioning explicitly on the study inclusion mechanism, the amortized approach yields parameter estimates that remain coherent across virus variants and selection mechanisms.

The trained bias-aware NPE was subsequently applied to the real PedCovid data. Across both the Alpha and Omicron variant cohorts, posterior estimates obtained from the bias-aware NPE and from the likelihood-based MCMC approach differed systematically, with the largest discrepancies observed in parameters governing age-related differences in infection risk ($\mu_{\text{inf}}$) and the household-size-dependent transmission parameter $\delta$ (\autoref{fig:pedcov_results}\refsubfigure{D}). In the Alpha variant, infection-related modifiers, particularly those associated with asymptomatic individuals, were substantially larger under the bias-aware approach, whereas the MCMC estimates shrank these effects toward values near or below one (\autoref{fig:pedcov_results}\refsubfigure{D}). Protection parameters also shifted qualitatively for both variants, with bias-aware inference generally implying weaker protection against transmission and smaller effects on acquisition relative to MCMC for the Alpha variant and stronger protection and acquisition effects for Omicron. These patterns suggest that ignoring the selection mechanism induces structural bias in several transmission parameters while leaving the baseline transmission rate largely unaffected.

Simulation-based calibration further supports this interpretation. Under MCMC, symptomatic infection-related modifiers were systematically underestimated, whereas susceptibility-related modifiers were on average overestimated across simulated datasets for the child selection procedure (\autoref{fig:pedcov_results}\refsubfigure{C}). 
On the real data, $\mu_{\text{inf}}^{\text{sym Infant}}$ and $\mu_{\text{inf}}^{\text{sym Child}}$ appear underestimated relative to the bias-aware NPE posterior (\autoref{fig:pedcov_results}\refsubfigure{D}), which aligns with the SBC direction.
Finally, classifier-based diagnostics confirm the consistency of the bias-aware posterior with the assumed generative model. For the real PedCovid data, the classifier accuracy was $0.719$ ($p=0.50$) for the Alpha variant and $0.766$ ($p=0.20$) for the Omicron variant, indicating no statistically significant discrepancy between posterior samples obtained with the bias-aware NPE and samples drawn from the joint simulator. Together, these results demonstrate that incorporating the selection mechanism directly into the simulator enables calibrated posterior inference even in complex stochastic models where explicit likelihood-based correction of selection bias is infeasible.

\section*{Discussion}

In this study, we developed a general amortized Bayesian inference framework that accounts for selection bias in parameter estimation tasks. By embedding selection, censoring, and missingness within forward simulations and training neural networks on the resulting joint distribution, the proposed approach yields calibrated and testable posterior inferences, even under complex selection mechanisms. Across three simulated and real epidemiological applications, the method matches analytic solutions where such solutions are available and remains applicable in regimes where the formulation of likelihoods accounting for the selection process is difficult.

A key strength of the framework is that it requires only a simulator of the population process together with the selection mechanism. In many application areas, such simulation models are routinely developed to study bias through Monte Carlo experiments, sensitivity analyses, or study design investigations \cite{burton2006design, rothman2008modern, kawabata2024accounting}. Our approach leverages this established modeling practice for inference rather than using simulations solely for validation or sensitivity analysis. Because the method operates entirely through forward simulation, it can accommodate arbitrary selection operators, including nonlinear, stochastic, and latent-state-dependent mechanisms, provided that they can be simulated. This flexibility allows the framework to be applied in settings where analytical likelihood corrections are unavailable or difficult to derive. More broadly, the growing role of simulation-based approaches in statistical workflows \cite{burkner2025simulations} highlights the potential of such methods to address complex data-generating processes that cannot be easily expressed in closed form.

An additional advantage of the framework is amortization. Once trained, the inference network can be applied to multiple datasets without retraining, enabling rapid posterior estimation across simulated scenarios and real-world studies in contrast to neural likelihood-based approaches \cite{boyd2024accounting}. This capability makes the approach particularly useful for systematically exploring the consequences of selection bias under a range of plausible assumptions. For example, once a simulator has been specified and an inference network trained, researchers can efficiently evaluate how alternative data-collection procedures or selection mechanisms affect inference without re-deriving likelihoods or repeating computationally expensive analyses. In this way, amortized simulation-based inference provides a practical tool not only for bias-aware parameter estimation but also for sensitivity analysis \cite{elsemueller2024sensitivityaware} and experimental design \cite{bracher2025jadai}.

A potential concern is misspecification of the selection mechanism. While the proposed framework requires specifying a structural form for the selection process, such misspecification does not remain silent: discrepancies between the training data and the observed data can be detected through the C2ST diagnostics.
Importantly, the selection mechanism need not be fully known. Unknown components, such as parameters governing prevalence or inclusion probabilities, can be incorporated into the generative model as additional latent variables and inferred jointly within the same amortized framework. In this way, uncertainty about the selection process itself can be propagated through the posterior, provided that a plausible structural form can be specified. This capability distinguishes the proposed approach from most existing methods addressing bias or missingness. Previous work has primarily focused on settings with partially observed covariates or missing data under missing-at-random assumptions \cite{lueckmann2017flexible, WangHas2024, gloeckler2024all, verma2025, simkus2025}, where the observation process is either assumed known or modeled through auxiliary imputation mechanisms. Other approaches target specific bias structures, such as confounding addressed through instrumental-variable assumptions \cite{braun2025flowivcounterfactualinference}, or aim to improve robustness to model misspecification \cite{bharti2026amortised, khoo2026minimum}. These methods typically rely on additional identification conditions or prespecified observation models and therefore do not allow uncertainty in the selection mechanism itself to be inferred jointly with the parameters of interest. By contrast, our framework treats the selection process as an explicit component of the generative model, allowing unknown aspects of the selection mechanism to be parameterized and marginalized over during inference. Consequently, the approach does not eliminate the need for substantive modeling assumptions; rather, it reframes selection bias from a problem of post hoc statistical correction to one of explicit generative modeling, where uncertainty about both the outcome and selection processes is propagated coherently through the posterior.

As with other neural posterior estimation methods, performance depends on the choice of network architecture, the available simulation budget, and the fidelity of the underlying simulator. Insufficient coverage of the parameter space or simulator misspecification may lead to biased or poorly calibrated posteriors \cite{SchmittBue23}. In addition, the classifier two-sample tests used for diagnostic evaluation operate on the low-dimensional embedding learned by the summary network rather than on the raw data. While such dimensionality reduction is necessary for computational tractability in high-dimensional settings, discrepancies that are not captured by the learned summary statistics may remain undetected. Consequently, both posterior quality and diagnostic reliability depend on the expressiveness of the neural networks and on how well the simulated training distribution reflects the relevant range of real-world data-generating processes.
In this context, well-behaved simulation-based calibration results provide empirical evidence that the chosen network architectures are sufficiently expressive for the problem at hand, given the available simulation budget.

In summary, we establish a simulation-based framework for Bayesian inference under selection bias that embeds the selection mechanism directly within the generative model. By combining amortized neural posterior estimation with explicit modeling of the data-collection process and formal diagnostic evaluation, the approach yields calibrated inference in settings where likelihood-based corrections are unavailable or impractical. 
We provide a reusable implementation directly embedded in state-of-the-art tools with well-documented examples.
As simulation-based methods become increasingly central in applications involving complex and structured data acquisition, this framework provides a principled basis for integrating selection mechanisms into routine probabilistic analysis.

\section*{Methods}

\subsection*{Problem Description}\label{sec:problem}

We consider posterior inference $p(\thetab \mid \Y, \S=1; \C)$ given the selected observations $\Y=\{\y_i\}_{i=1}^N$ with $\S{=}1$ and the corresponding covariates $\C=\{\c_i\}_{i=1}^N$. 
Selection is considered a missing data problem \cite{howe2015all}.
We write $\Y^{\text{full}}=(\Y, \Y^{\text{mis}})$, where $\Y$ denotes the observed values and $\Y^{\text{mis}}$ denotes the missing values (due to selection).
Then the likelihood has the form
\begin{align}\label{eq:full_likelihood}
p(\Y, \S \mid \thetab; \C) 
= \iint &
p(\Y, \Y^{\text{mis}} \mid \thetab; \C, \C^{\text{mis}})
\, p(\S \mid \Y, \Y^{\text{mis}}, \thetab; \C, \C^{\text{mis}}) \d\C^{\text{mis}}\d\Y^{\text{mis}},
\end{align} 
where $p(\S \mid \Y, \Y^{\text{mis}},\thetab; \C, \C^{\text{mis}})$ reflects the probability of a sample being selected during the selection process.
In addition, the selection probability might depend on the latent variables, which can be computed from model parameters and covariates.
To get the unbiased posterior distribution
\begin{equation}
p(\thetab \mid \Y, \S; \C) \propto p(\thetab) p(\Y, \S \mid \thetab; \C),
\end{equation}
all unobserved variables must be marginalized.
For a detailed introduction to selection bias, we refer to \cite{little2019statistical}.

We differentiate between three scenarios according to the respective missingness type.

First, if the selection probability $p(\S \mid \Y, \Y^{\text{mis}},\thetab; \C, \C^{\text{mis}})$ depends on neither $\Y$ or $\Y^\text{mis}$ nor the covariates $\C$ or $\C^{\text{mis}}$, then this is called \emph{missing completely at random}.
This is the simplest case, because missingness can be ignored.

Second, if the selection process depends only on $\Y$ or $\C$ and not on $\Y^\text{mis}$ and $\C^{\text{mis}}$, the integrals in \eqref{eq:full_likelihood} simplify to 
$p(\Y, \S \mid \thetab; \C) 
=  p(\Y \mid \thetab; \C) p(\S \mid \Y,\thetab; \C)$.
This scenario is called \emph{missing at random}.
Furthermore, if $\thetab=(\boldsymbol{\phi},\boldsymbol{\psi})$ and $p(\Y, \S \mid \thetab; \C)=p(\Y \mid \boldsymbol{\phi}; \C) p(\S \mid \Y,\boldsymbol{\psi}; \C)$, and the prior factorizes as $p(\thetab)=p(\boldsymbol{\phi}) \, p(\boldsymbol{\psi})$, then the missing-data mechanism is ignorable for Bayesian inference \cite{little2019statistical}.
However, inferred population-level quantities are only valid for the observed covariates.
If the target of inference is a population-level parameter, such as the prevalence of infection in the full population rather than in the selected sample $\Y$, then the selection depending on $\C$ can no longer be ignored.
In this case, the observed covariate distribution $p(\C)$ differs from the population distribution $p(\C^{\text{full}})$ and must be explicitly corrected.

Third, when the selection mechanism depends on unobserved values such as $\Y^{\text{mis}}$ and $\C^{\text{mis}}$ the data are \emph{not missing at random}. In this case, the selection process cannot be ignored for inference, and all unobserved quantities involved in the selection mechanism must be explicitly modeled and marginalized to obtain unbiased posterior inference. 

Failure to account for the dependence of selection on unobserved variables generally leads to biased parameter estimates and invalid uncertainty quantification~\cite{little2019statistical}.

\subsection*{Amortized Bayesian Inference Under Selection Bias}\label{sec:abi}

We train a neural posterior estimator (NPE) to learn a global mapping from datasets to posterior distributions across repeated simulations from a generative model. Let $\thetab \sim p(\thetab)$ denote parameters drawn from a prior distribution, and $\Y \sim p(\Y \mid \thetab; \C)$ denote data generated from the forward model. 
We sample $\C$ from the data, resulting in an implicit prior.
We simulate parameter-dataset pairs $(\thetab, \Y; \C)$ from the joint distribution $p(\thetab, \Y, \C)$ and train a NPE $q(\thetab \mid \Y; \C)$ to approximate the true posterior $p(\thetab \mid \Y; \C)$. 
Here, we instantiate the NPE using two neural networks: a summary network and a generative neural network as an inference network, for example, a flow matching or consistency model (see \cite{arruda2025diffusionSBI} for a detailed introduction to generative neural networks and the corresponding objective functions). 
Both networks can be trained jointly, where the summary network provides a lower-dimensional summary of the simulation to the inference network.
Under sufficient model capacity and training data, this procedure yields a consistent approximation of the posterior \cite{RadevMer2020bayesflow}.

In the presence of selection bias, we augment the generative model with an explicit selection mechanism. Parameter-dataset pairs are generated as before from the structural population model, yielding a latent population-level dataset. A selection mechanism is then applied to this dataset, resulting in a random variable $\S$, representing sampling or missingness processes that determine which units are observed. In this way, selection acts on the fully simulated population data prior to observation. Because the framework relies only on forward simulation, it accommodates arbitrary and potentially complex selection mechanisms provided that they can be simulated.

The resulting estimator approximates $p(\thetab \mid \Y, \S=1; \C)$ without requiring the evaluation of a tractable likelihood or explicit estimation of selection probabilities. 
Moreover, we train the NPE across multiple selection mechanisms by parameterizing or indexing the selection process and conditioning the network on the corresponding selection indicator or mechanism-specific variables.
The specific NPE architecture used in each experiment is described in the following sections.

\subsection*{Simulation-Based Diagnostics for Bias Detection}\label{sec:diag}

To assess the calibration of the approximate posterior distributions, we employed simulation-based calibration (SBC) \cite{TaltsBet2020}. SBC evaluates whether posterior uncertainty is correctly quantified by repeatedly simulating parameters and datasets from the prior and generative models, performing inference, and computing the rank of each true parameter value within its corresponding posterior sample. 
If the posterior is calibrated, these ranks are uniformly distributed across repeated simulations. 

In practice, we generated $M$ independent parameter draws $\thetab^{(m)} \sim p(\thetab)$, simulated the corresponding datasets $\Y^{(m)}$, and obtained posterior samples $\tilde{\thetab}^{(m)}_k$ from the trained NPE (or any other inference algorithm).
For each parameter component, we computed the rank of $\thetab^{(m)}$ among the $K$ posterior samples. Deviations from uniformity in the empirical ranks indicate systematic bias, overconfidence, or underdispersion in the posterior approximation.
This deviation can be graphically assessed by computing the empirical cumulative distribution function (ECDF) and comparing it with the known uniform CDF \cite{TaltsBet2020}.
In selection settings, SBC is performed under the full generative model, including the selection mechanism, allowing controlled quantification of bias across different selection regimes.

As an additional diagnostic, we used a classifier two-sample test (C2ST) \cite{lopez-paz2017revisiting} to assess the fidelity of the learned posterior approximation. C2ST evaluates whether samples drawn from the learned posterior, combined with simulations, are statistically indistinguishable from samples drawn directly from the true joint distribution \cite{yao2023discriminative, linhart2023c2st}.

Specifically, we constructed two sets of samples: (i) samples obtained by drawing a single $\thetab \sim q(\thetab \mid \Y)$ for each dataset and pairing it with the corresponding $\Y$ (ii) joint samples $(\thetab, \Y)$ generated from the prior and forward models, including the selection mechanism. 
A binary classifier was then trained to distinguish between these two sample sets. 
If the posterior approximation is accurate, the classifier should achieve chance-level performance on the held-out data. In practice, classifiers were instantiated as a neural network and trained using 5-fold cross-validation on validation data not used for training the NPE.

Classification accuracy substantially above chance indicates discrepancies between the learned posterior-data pair and the true joint distribution, suggesting insufficient training of the NPE or remaining bias. 
To assess the statistical significance for a given observed dataset, we additionally trained $B{=}10$ classifiers on label-permuted data, following established C2ST procedures \cite{linhart2023c2st}.
We define the test statistic as
\begin{equation}
    T = \frac{1}{K} \sum_{k=1}^{K} \left(a_k - \frac{1}{2}\right)^2,
\end{equation}
where $a_k$ denotes the classification accuracy for the parameter-data pair $k$.
The empirical distribution of test statistics $T_b$ under permutations $b=1,\dots,B$ provides a reference against which the observed test statistic $T_{\text{obs}}$ is compared, yielding a $p$-value:
\begin{equation}
    p = \frac{1}{B} \left( \sum_{b=1}^{B} \mathbb{I}(T_b \geq T_{\text{obs}}) \right).
\end{equation}
Beyond validation on simulated data, C2ST can be applied to real datasets by applying the classifier to observed data paired with posterior samples.
This enables the detection of bias or violations of modeling assumptions even when the true parameters are unknown.
To simplify the training of the classifiers, we first projected the data $\Y$ onto a lower-dimensional embedding using the summary network of the NPE.
The classifiers then consisted of two hidden multi-layer perceptrons  with widths such that the width was larger than 10 times the input dimension, as implemented in the \texttt{BayesFlow} library \cite{kuehmichel2026bayesflow2}.

\subsection*{Estimating Prevalence Under Selection Bias}\label{sec:prevalence}
We consider estimation of the population-level prevalence $\rho = \E[\Y^{\text{full}}]$ under the selection $p(\S \mid \C^{\text{full}})$ motivated by the KoCo19 Study \cite{radon2020protocol}.
The simulation framework incorporates covariate-dependent infection risk, non-representative sampling, outcome and covariate missingness, and imperfect diagnostic testing.

\subsubsection*{Logistic Infection Model and Population Construction}
Individual-level infection status was modeled using logistic regression with odds ratio parameterization. The linear predictors included sex, age group, country of birth, and household size, with reference categories corresponding to male sex, age 20--34 years, birth in Germany, and single-person households. 
The model parameters consisted of an intercept on the log-odds scale and log-odds ratios for each non-reference category:
\begin{equation}
    \tilde{y}_i \sim \mathrm{Bernoulli}(p_i), \qquad \text{logit}(p_i)=\beta_0+\boldsymbol{\beta}^\top \c_i.
\end{equation}
For simulation-based experiments, parameters were drawn independently from Gaussian prior distributions on $\beta_0$ with mean $-3$ and standard deviation of $1.0$ and on $\boldsymbol{\beta}$ with mean $0$ and standard deviation of $0.5$. The prior was chosen to be mostly uninformative and to obtain simulated prevalence below $50\%$ of the population. The observed test results were then generated from $\tilde{y}_i$ using a misclassification model parameterized by externally estimated test sensitivity ($\mathrm{Se}{=}0.886$) and specificity ($\mathrm{Sp}{=}0.997$) \cite{olbrich2021head}, yielding apparent infection status $y_i$.

To construct latent population-level data, observed KoCo19 participants were oversampled using iterative proportional fitting, that is, $p(\S \mid \C)$ was estimated to match known marginal population distributions of Munich from \cite{radon2020protocol} with respect to age, sex, household size, and country of birth. 
For simulation experiments, we generated a synthetic population equal to $10\%$ of the Munich population; for the real data analysis, we generated the full 1.5 million inhabitants.
Missing covariates were imputed during this step by sampling from the corresponding population-level target distributions, as they were assumed to be missing completely at random. 
The resulting oversampled dataset approximated a synthetic population consistent with known census margins, where true infection outcomes were then simulated using the above logistic model and prevalence calculated as $\rho=\E[\Y^\text{full}]$. 

Outcome missingness was imposed by reproducing the missingness pattern observed in the original KoCo19 data. Hence, individuals with missing outcomes in the original study also had missing outcomes after oversampling, thereby preserving the empirical missing at random structure.
To emulate the original study design, the synthetic population was subsequently downsampled to the original cohort size using inverse-probability subsampling based on oversampling weights, yielding a biased sample that matched the observed KoCo19 cohort. This procedure preserved both the selection bias and missingness structure of the real data.

\subsubsection*{Prevalence Estimation}
Two baseline prevalence estimators were considered. First, an unadjusted estimator based on complete cases was computed from the apparent test outcomes of the (simulated or real) KoCo19 cohort and corrected for test misclassification using the Rogan-Gladen estimator \cite{RoganGla1978}:
\begin{equation}
\hat{\rho}_{\text{RG}} 
= \frac{\hat{\rho}_{\text{obs}} + \mathrm{Sp} - 1}{\mathrm{Se} + \mathrm{Sp} - 1}.
\end{equation}
Second, inverse probability weighting was applied to account for unequal sampling probabilities induced by the subsampling step, followed by misclassification correction. 
For bootstrap-based uncertainty quantification \cite{efron1994introduction}, the entire pipeline was repeated with 100 bootstrap resamples of the original KoCo19 data. 
In the analyses of the real data, simulated outcomes were replaced by observed test results, while the same inference procedures were applied unchanged.

\subsubsection*{Neural Posterior Estimation}

To perform amortized Bayesian inference, we used a NPE with a simulation-based workflow tailored to the KoCo19 cohort.
We used the simulation pipeline described above to generate training data consisting of $J=10{,}000$ pairs $\{(\Y_j, \rho_j)\}_{j=1}^J$ of subsampled cohorts $\Y_j$ with a known prevalence $\rho_j$ and 1000 additional validation pairs.
Here, we aim to directly estimate the population-level prevalence and not the underlying parameters of the logistic model, which we marginalize implicitly.
The observation $\Y_j$ includes the observed test results $\y_{j,i}$ and covariates $\c_{j,i}$ for all individuals $i\in\{1,\ldots,5577\}$.
As all covariates $\c_{i}$ were categorical, we converted them to non-negative integers.
Missing values were encoded using a fixed value of $-1$, following the approach of \cite{WangHas2024}, allowing the network to learn to distinguish between observed and unobserved inputs.

The workflow then proceeded by training a NPE, consisting of an inference network and a summary network, which were trained jointly using the training data.
The input to the summary neural network consisted of $\Y_j$ and the index of the epoch $e_j\in\{1,2,3,4,5\}$ used as the basis for oversampling, which determined the distribution of the covariates and the amount of missingness.
The summary network was instantiated as a deep set encoder that can learn permutation-invariant representations of set-based data \cite{zaheer2017deep}, with a summary dimension of 4.
This choice ensured that the summary network was agnostic to the order of individuals in the cohort.
For the inference network, we used a small conditional flow matching model \cite{wildberger2023flow}, as it also allows for posterior density estimation, with 2 conditional multi-layer perceptrons of width 128 as the backbone and a dropout rate of $0.1$.
For an in-depth introduction to SBI with generative models, refer to~\cite{arruda2025diffusionSBI}.
For training, we used the inference pipeline implemented in the \texttt{BayesFlow 2.0.8} library~\cite{kuehmichel2026bayesflow2} with \texttt{jax} as the backend, utilizing the default training settings.
We trained for 300 epochs using the AdamW optimizer \cite{loshchilov2018decoupled}, with a batch size of 64, an initial learning rate of $5{\times}10^{-4}$, and a cosine schedule for the learning rate \cite{loshchilov2017sgdr}.
Furthermore, the prevalence was constrained between 0 and 1 by applying a sigmoid transformation before training.

To evaluate the different inference procedures, we computed the absolute difference between the estimated prevalence and the true prevalence using 1000 newly generated simulations.
With the NPE, inference is amortized; hence, after training the model once, we sampled 500 posterior samples and additionally evaluated the posterior density for each sample to approximate the mode of the posterior.
We then used the mode as a point estimator of the prevalence for comparison with other baseline estimators.
Furthermore, we assessed the reliability of the approximate posteriors produced by the NPE using C2ST.

\subsection*{Estimating Time-to-Event Under Missing Disease due to Death Bias}\label{sec:cens_visit}

We consider time-to-event analysis in a semi-competing risks setting in which death may preclude the observation of dementia onset, as in incidence studies based on the Framingham Heart Study \cite{satizabal2016incidence,binder_letter_2016}. 
Let $\Y$ denote the time-to-dementia or death outcome and $\S$ the selection indicator induced by unobserved dementia onset before death. 
The target is the posterior $p(\thetab \mid \Y,\S; \C)$ where censoring induces the selection probability $p(\S \mid \Y, \Y^{\text{mis}}, \thetab, \C, \C^{\text{mis}})$, i.e., not missing at random.
This structure is embedded in a SBI framework.

\subsubsection*{Illness-Death Multi-State Model}

Event dynamics were modeled using an illness-death multi-state model with three states: healthy (0), dementia (1), and death (2). Transitions between states followed cause-specific hazard models for the transitions $(0\to1)$ (dementia onset), $(0\to2)$ (death without prior dementia), and $(1\to2)$ (death after dementia), following \cite{joly2002penalized, binder_estimating_2017, binder2019multi}.
Baseline hazards ($h_{01}$, $h_{02}$, $h_{12}$) were parameterized by transition-specific scale parameters $a_{kl}$ ($k\to l$) and shape parameters $\kappa_{kl}$ governing Weibull hazard functions.
The covariate effects $\boldsymbol{\beta}_{kl}$ of sex and age were included as transition-specific log-linear modifiers of the transition hazards:
\begin{equation}
h_{kl}(t \mid \c)=
a_{kl}\,\kappa_{kl}\, t^{\kappa_{kl}-1}  \,\exp(\boldsymbol{\beta}_{kl}^\top \c).
\end{equation}
Age was centered before inclusion in the linear predictors.

For simulation-based experiments, the transition parameters were drawn independently from the prior distributions. Baseline scales $a_{kl}$ and shape parameters $\kappa_{kl}$ were assigned Gamma priors, parameterized through their mean $m$ and coefficient of variation $v$. Specifically, $a_{kl} \sim \mathrm{Gamma}(m_{a}=0.0002993, v_{a}=1)$ and $\kappa_{kl} \sim \mathrm{Gamma}(m_{\kappa}=1, v_{\kappa}=0.25)$, corresponding to moderate deviations from exponential hazards. 
Covariate effects for sex and age were assigned independent Gaussian priors $\N(0, 1^2)$.
This prior specification induced realistic heterogeneity in event times while remaining weakly informative (\autoref{fig:appendix_cens_vist_recovery}).

Given the sampled parameters and individual-level covariates, complete event trajectories were generated by inverse transform sampling from cause-specific hazard functions. For each individual, competing event times for dementia onset and direct death were simulated, with the earliest event determining the first transition. An additional transition time from dementia to death was simulated for individuals who experienced dementia onset.

\subsubsection*{Visit-Based Censoring and Observation Process}
The Framingham data were obtained from the Framingham Heart Study, a long-term cohort study initiated in 1948. It comprises an original cohort of 5209 residents of Framingham, Massachusetts, with repeated examinations every two years, and an offspring cohort of 5214 participants initiated in 1971 with examinations approximately every four years. Following \cite{satizabal2016incidence} and \cite{binder2019multi}, both cohorts were combined for analysis, resulting in comparable sample sizes across four non-overlapping epochs, each spanning a five-year period.
To reflect the structure of longitudinal cohort studies, simulated event times were transformed into observed data through a visit-based censoring mechanism extending \cite{binder2019multi}. 
Each individual was assigned two observation times corresponding to the study visits. 
Dementia status was only observable at visit times, inducing censoring of dementia onset.
If death occurred before dementia was detected at a visit, the dementia status was censored to the last known status.
If dementia occurred after the last visit, the dementia status was censored.
Administrative censoring was applied at the end of the follow-up.
If death or dementia occurred after the study ended, the corresponding event was censored. 
If the patients were disease-free at the end of the epoch and continued in the next epoch, the last visit time point was set to the end of the epoch.
In addition, to reflect random dropout, healthy individuals were assumed to have a $5\%$ probability of leaving the study, independent of covariates and event times. 
The mean visit times were estimated empirically for each study epoch. For each individual, two visit times were then generated by sampling around these epoch-specific means: the first visit time, corresponding to observations within the first 2.5 years of follow-up, was drawn with variance 100, and the second visit time, corresponding to observations after 2.5 years, was drawn with variance 150. 
Simulations were conducted separately for each of the four Framingham Heart Study epochs. For each epoch, individual-level covariates were obtained directly from empirical data. 
For each simulated dataset, the model parameters were sampled from the priors.
This procedure resulted in simulated data that closely mimicked the visit structure of the Framingham Heart Study.

\subsubsection*{Neural Posterior Estimation}

Simulated visit-censored datasets were used to train a neural posterior estimator within a SBI framework. 
We generated 20,000 training datasets consisting of pairs of censored cohorts $\Y_j$ and the 12 corresponding parameters $\thetab=(a_{kl},\kappa_{kl},\boldsymbol{\beta}_{kl})$ of the multi-state model.
Each individual $\y_{j,i}\in\Y_j$ was then described as a vector of the time of illness, binary indicator of illness, time of death, binary indicator of death (all possibly censored), sex, age, and epoch ID.
We standardized the covariates and time variables before training and log-transformed the scale and shape parameters. $\Y_j$ was padded with $-1$, such that it had the same number of individuals (2299) for every epoch.

The summary network of the NPE was instantiated as a set transformer to learn permutation-invariant representations \cite{lee2019set} with a summary dimension twice the number of parameters.
As an inference network, we used a consistency model due to its inference speed \cite{schmitt2024consistency}.
The backbone of the consistency model was chosen as the default 5 conditional multi-layer perceptrons of width 256 with a dropout rate of 0.1.
We trained the model using the same settings as before in the \texttt{BayesFlow} library \cite{kuehmichel2026bayesflow2}.
Similar to the previous model, we assessed the reliability of the approximate posteriors produced by the NPE using the C2ST-diagnostic on a validation set with 1000 cohorts.

We compared our approach to a full likelihood approach and a naive Cox model \cite{joly2002penalized, binder2019multi}.
For the Cox approach, separate models were fitted for each transition ($0\to1$, $0\to2$, $1\to2$), with competing events treated as independent censoring. In contrast, the likelihood-based approach relies on an illness-death model, where all observed cases of transition, including missing disease by death, are handled through exact likelihood contributions derived from the multi-state process, while penalization enforces smoothness of the hazard functions \cite{joly2002penalized}.
Baseline hazards were parameterized by splines, and age- and sex-adjusted cumulative hazards were obtained by evaluating the linear predictor at the mean age and sex values, multiplying the baseline hazards by the resulting proportional hazards factor, and integrating the adjusted hazards over time following \cite{binder2019multi}.

In addition, a NPE with identical architecture and training configuration was trained on uncensored data subject only to administrative censoring. 
Posterior calibration was assessed using C2ST on the corresponding validation set.
Hence, the second NPE and corresponding classifier were trained exclusively on uncensored data.

\subsection*{Estimating Transmission Rates Under Selection}\label{sec:pedcov}

We consider a stochastic household transmission model (described in detail in forthcoming work) with outcome-dependent study inclusion motivated by the PedCovid Study \cite{delaunaymoisan2022}. 
Let $\Y$ denote the observed household-level infection data and $\S$ the study inclusion indicator. 
Inference targets the posterior $p(\thetab \mid \Y,\S;\C)$ under a selection mechanism 
$p(\S \mid \Y, \Y^{\text{mis}}, \thetab, \C, \C^{\text{mis}})$,
that depends on unobserved infection outcomes and thus corresponds to a not missing at random setting.

\subsubsection*{Model Structure and Transmission Hazard}

Two SARS-CoV-2 variants (Alpha and Omicron) were separately modeled using variant-specific parameterizations. Let $\tau_i$ denote the infection time of individual $i \in \{1,\ldots,N\}$, with $\tau_i=\infty$ if individual $i$ is never infected. At time $t$, a susceptible individual $i$ experiences an infection hazard
\begin{equation}
\lambda_i(t)
=
\alpha
+
\beta\, w(n)
\sum_{j:\,\tau_j < t}
\kappa(t-\tau_j)\,
\mu_{\text{inf}}
\mu_{\text{sus}}
\mu_{\text{pro}}
\end{equation}
from other household members $j$, where $\alpha$ is a background hazard, $\beta$ is the baseline transmission intensity, and $w(n)=(n/4)^{-\delta}$ scales the transmission by household size $n$ with exponent $\delta$. The generation-time kernel $\kappa(\cdot)$ was modeled as a Gamma density, with parameters chosen separately for the Alpha variant ($\text{shape}{=}2$, $\text{rate}{=}0.44$) \cite{chen2022generation} and Omicron variant ($\text{shape}{=}3.351$, $\text{rate}{=}1.1098$) \cite{anderHeiden_Buchholz_2022}.

The multiplicative terms $\mu_{\text{inf}}$, $\mu_{\text{sus}}$, and $\mu_{\text{pro}}$ encode heterogeneity in infectiousness, susceptibility, and protection by vaccination or past exposures. Specifically, $\mu_{\text{inf}}$ denotes the infectivity multiplier of an infectious individual $j$ at time $t$, defined relative to symptomatic adults and varying by age category (infant ${<}6$ years, child 6--11 years, and adult ${>}11$ years) and symptom status (symptomatic or asymptomatic). The term $\mu_{\text{sus}}$ represents the susceptibility multiplier of a susceptible individual $i$, defined relative to adults and depending on age, while $\mu_{\text{pro}}$ captures the protection effects on transmission or acquisition.

Priors were specified as weakly informative while reflecting plausible transmission dynamics. The baseline transmission intensity $\beta$ was assigned a Gamma prior with shape 2.0 and scale 0.5, and the household size scaling exponent $\delta$ was assigned a standard normal prior. All multiplicative effects capturing infectiousness, susceptibility, and protection were assigned log-normal priors with a log-mean of 0 and log-standard deviation of 0.7.
The background hazard $\alpha$ was fixed to 0.001 for the Alpha variant and 0.01 for the Omicron variant, due to weak identifiability of this parameter under random household selection (\autoref{fig:appendix_pedcvo_recovery}).

Over a short interval $[t,t+\Delta t)$, the probability that an individual $i$ becomes infected is
\begin{equation}
\Pr\!\left(i \text{ infected in } [t,t+\Delta t)\right)
=
1-\exp\!\left(-\lambda_i(t)\Delta t\right).
\end{equation}

Following infection, individuals were assigned as asymptomatic, with a probability of 0.4 for the Alpha variant and 0.3 for the Omicron variant. For symptomatic individuals, the incubation period was modeled by a Gamma distribution, with parameters differing by variant (Alpha: $\text{mean}{=}\SI{4.42}{d}$, $\text{SD}{=}\SI{2.30}{d}$; Omicron: $\text{mean}{=}\SI{3.09}{d}$, $\text{SD}{=}\SI{1.64}{d}$) \cite{GALMICHE2023e409}. The symptom onset times are given by $D_i=\tau_i+\text{incubation}_i$.

Symptomatic individuals underwent testing after symptom onset, with testing delays drawn from a geometric distribution with a success probability of 0.33 for both variants. A test was assumed to yield a positive result between 1 and 15 days after infection. 
A positive test by a symptomatic individual triggered the testing of other household members with household-size-specific probabilities of missing tests, estimated empirically from the PedCovid data (\autoref{tab:missing_test}). 
Triggered tests occurred after delays were drawn from geometric distributions (Alpha: $p{=}0.48$, Omicron: $p{=}0.46$).
In addition, all individuals were subjected to background testing for other reasons, with fixed daily testing probabilities of $1/21$ for Alpha and $1/14$ for Omicron.
For asymptomatic individuals, the observation time was defined as the time of the first positive test.

The transmission dynamics were simulated using a fixed-step tau-leaping scheme with daily time steps. Household covariates were obtained directly from the PedCovid dataset, and each household was simulated 50 times.  
At each time step, susceptible and infectious individuals were identified, infection hazards were computed, new infections were sampled, and symptom status and incubation periods were assigned as appropriate. Testing events, inclusion procedures, and follow-up testing were then applied, after which households were selected according to household selection schemes.

\begin{table}[t]
\centering
\caption{Probability of a household member missing a test triggered by a symptomatic member testing positive estimated from the PedCovid data.}
\label{tab:missing_test}
\begin{tabular}{cccccccc}
\toprule
\diagbox{Variant}{Household size} & 2 & 3 & 4 & 5 & 6 & 7 & 8 \\
\midrule
Alpha   & 0.00 & 0.10 & 0.07 & 0.14 & 0.10 & 0.71 & -- \\
Omicron & 0.00 & 0.40 & 0.24 & 0.17 & 0.17 & 0.14 & 0.13 \\
\bottomrule
\end{tabular}
\end{table}

\subsubsection*{Household Selection Schemes}

Household inclusion in the study occurred once a household member tested positive. From all individuals with a positive test result within the age range defined by the selection scheme, a single inclusion case was randomly chosen. The household inclusion date was set to $D_k + \text{delay}$, where the delay followed a Poisson distribution with a mean of 4.8 days for Alpha and 2.7 days for Omicron. Follow-up tests in the household for all members were scheduled at 3, 7, 15, and 45 days after inclusion. All dates were shifted by 30 days to match the earliest infected individuals in the data.
All geometric and Poisson delay distributions were fitted by maximum likelihood to the PedCovid data in the original study.

From the full set of simulated households, subsets were selected using different selection mechanisms. The target sample sizes were $N{=}128$ households for the Alpha variant cohort and $N{=}54$ for the Omicron variant cohort. Under random selection, households were sampled uniformly without replacement. Under outcome-dependent selection, households were included only if the designated inclusion case met a specified age criterion (child: ${<}18$ years; adult: ${\geq}18$ years). 
The child-based selection scheme replicates the procedure used in the PedCovid Study \cite{delaunaymoisan2022}.

\subsubsection*{Likelihood for MCMC and Random Sampling}

Given the observed infection and testing histories, the likelihood of the household transmission model can be defined by combining the infection hazard, survival contributions for uninfected individuals, and incubation period densities for symptomatic cases under the assumption of random selection of households.
The cumulative hazard for an individual $i$ up to time $t$ was
\begin{equation}
\Lambda_i(t)
=
\alpha(t-\tau_{\text{first},i})
+
\beta\, w(n)
\sum_{j:\,\tau_j<t}
F_{\text{lag}}(t-\tau_j)\,
\mu_{\text{inf}}
\mu_{\text{sus}}
\mu_{\text{pro}},
\end{equation}
where $\tau_{\text{first},i}$ denotes the first infection time in the household of the individual $i$, and
\begin{equation}
    F_{\text{lag}}(x)=\int_0^x \kappa(u)\d u.
\end{equation}
Unobserved infection times were treated as latent variables and constrained by observed testing data with soft quadratic penalties enforcing plausible temporal bounds.
Using this likelihood, inference is possible with MCMC under the assumption of random selection.
Specifically, we used the No-U-Turn Sampler as implemented in \texttt{Stan} \cite{carpenter2017stan} with 4 chains and 5000 samples after the burn-in.

\subsubsection*{Neural Posterior Estimation}

Household datasets were used to train a neural posterior estimator within a SBI framework. 
We generated 32,000 simulated household datasets $\{(\Y_j,\thetab_j)\}_{j=1}^J$, where each $\Y_j$ consisted of all observed testing and infection information for the households selected under a given selection mechanism, and $\thetab_j=(\mu_{\text{inf}}, \mu_{\text{sus}},\mu_{\text{pro}},\delta,\beta)$ denotes the 11 corresponding transmission model parameters.
Each individual in a household was represented by the features of age group, infection status (one-hot encoded), protection status, household size, virus variant, and selection mechanism (one-hot encoded), as well as the date of symptom onset (or the date of the first positive test for asymptomatic individuals), the date of the last negative test, the date of the first or last positive test, and the end of follow-up.
To enable batched training, households were padded to a maximum size of eight members using fixed padding values of $-1$.

Posterior inference was performed using a hybrid summary network similar to \cite{arruda2026simulation} combining 2 set transformers \cite{lee2019set}; one to capture within-household temporal structure and generate a latent representation of a household, and another one applied then on the set of representations to keep the permutation invariance property of households. 
The resulting summaries of size 3 times the number of parameters were passed to a flow matching model \cite{wildberger2023flow} with the default 5 conditional multi-layer perceptrons of width 256 with a dropout rate of 0.1 as backbone.
The NPE was trained for 150 epochs on the simulated datasets with a batch size of 64 using the \texttt{BayesFlow} library \cite{kuehmichel2026bayesflow2}.

Inference performance was evaluated using 320 simulated datasets for each of the 3 selection procedures. Posterior inference obtained via the neural posterior estimator was compared with MCMC, which relies on an explicit evaluation of the biased model likelihood.
Calibration of posterior distributions was assessed using SBC and C2ST. 

\subsection*{Code Availability}
Code is available at \href{https://github.com/arrjon/AmortizedSelectionBias}{github.com/arrjon/AmortizedSelectionBias}. 

\section*{Acknowledgments}
J.H. acknowledges financial support by the Deutsche Forschungsgemeinschaft (DFG, German Research Foundation) under Germany’s Excellence Strategy (EXC 2047 - 390685813, EXC 2151 - 390873048), by the European Union via ERC grant INTEGRATE (grant no 101126146), and by the University of Bonn (via the Schlegel Professorship of J.H.). 
N.B. acknowledges financial support by the Deutsche Forschungsgemeinschaft (DFG, German Research Foundation)  – Project-ID 499552394 – SFB 1597. 
S.C. acknowledges financial support from Université de Versailles Saint-Quentin-en-Yvelines, the Inception program (Investissement d’Avenir grant ANR-16-CONV-0005), Institut Pasteur, and the HOME project (ANR 20-CE35-0016).
The PedCovid study was funded by a grant from the French Ministry of Health (PHRC) and a grant from the ANR (RA-COVID-19). The authors thank the PedCovid working group, including Sylvie Behillil, Naïm Bouazza, Nelly Briand, Agnès Delaunay-Moisan, Flora Donati, Vincent Enouf, Jérémie Guedj, Marianne Leruez-Ville, Lulla Opatowski, Faheemah Padavia, Isabelle Sermet-Gaudelus, Chloé Sturmach, Sylvie van der Werf, and the technical team of the National Reference Center for Respiratory viruses.
We acknowledge the Marvin and Unicorn clusters hosted by the University of Bonn.
We thank Simon Cauchemez for helpful discussions on bias in household studies.

\section*{Author CRediT}
J.A.: Conceptualization, Formal analysis, Methodology, Software, Validation, Visualization, Writing – original draft, Writing – review \& editing.
S.C.: Data curation, Methodology, Software, Writing – review \& editing.
P.S.: Data curation, Formal analysis, Software, Writing – review \& editing.
A.W.: Investigation, Writing – review \& editing.
M.H.: Investigation, Writing – review \& editing.
I.S.: Investigation, Writing – review \& editing.
N.B.: Data curation, Methodology, Resources, Software, Supervision, Writing – review \& editing.
L.O.: Conceptualization, Methodology, Supervision, Writing – review \& editing.
J.H.: Conceptualization, Methodology, Funding acquisition, Project administration, Resources, Supervision, Writing – review \& editing.

\newcommand{\bibcommenthead}{} 
\bibliographystyle{apalike}
\bibliography{Database, Database2}

\clearpage
\appendix
\renewcommand{\thefigure}{S\arabic{figure}}
\setcounter{figure}{0}
\renewcommand{\thetable}{S\arabic{table}}
\setcounter{table}{0}

\section*{Supplementary Material}

\subsection*{Additional Results}
\begin{figure}[ht!]
\centering
\begin{subfigure}{\linewidth}   
    \centering
    \begin{overpic}[width=0.7\linewidth]{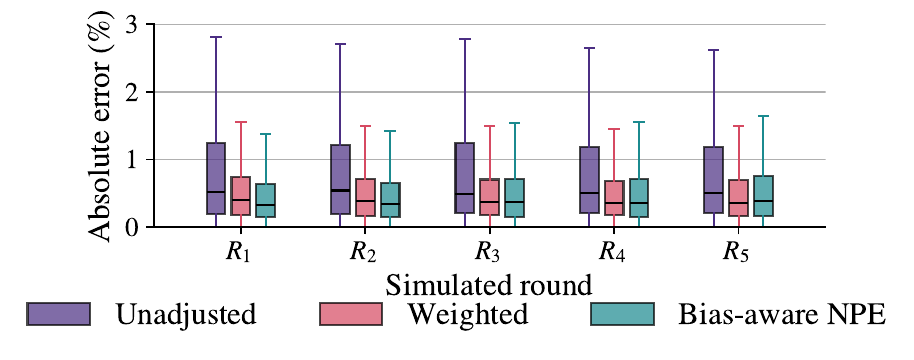}
        \put(-68,120){\textbf{A}}
    \end{overpic}
\end{subfigure}
\vfill
\begin{subfigure}{\linewidth}   
    \centering
    \begin{overpic}[width=\linewidth]{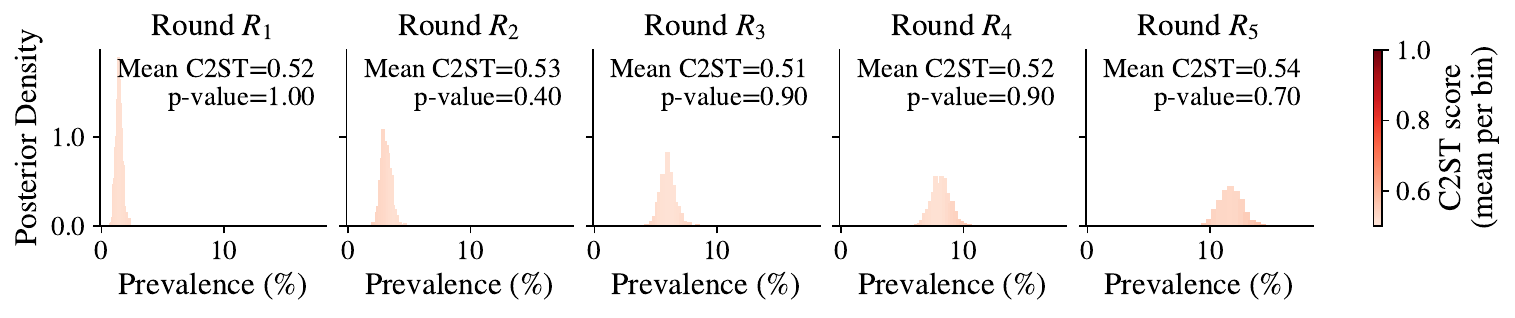}
        \put(0,90){\textbf{B}}
    \end{overpic}
\end{subfigure}
\caption{
    \emph{Additional results for the KoCo19 Study.}
    (\textbf{A}) Absolute error of estimated prevalence across 1000 simulated datasets (including selection, but excluding missingness), each with five rounds of the KoCo19 Study using unadjusted counts of infections and inverse-probability weighting. The mode of the posterior is estimated with neural posterior estimation on the dataset, including missingness.
    (\textbf{B}) Classification of the posterior marginals for the real data. A classifier was trained to distinguish samples from the joint $p(\thetab, \y)$ and $p(\thetab \mid \y)p(\y)$.
    }
\label{fig:prevalence_appendix}
\end{figure}

\begin{figure}[ht!]
\centering
\includegraphics[width=\linewidth]{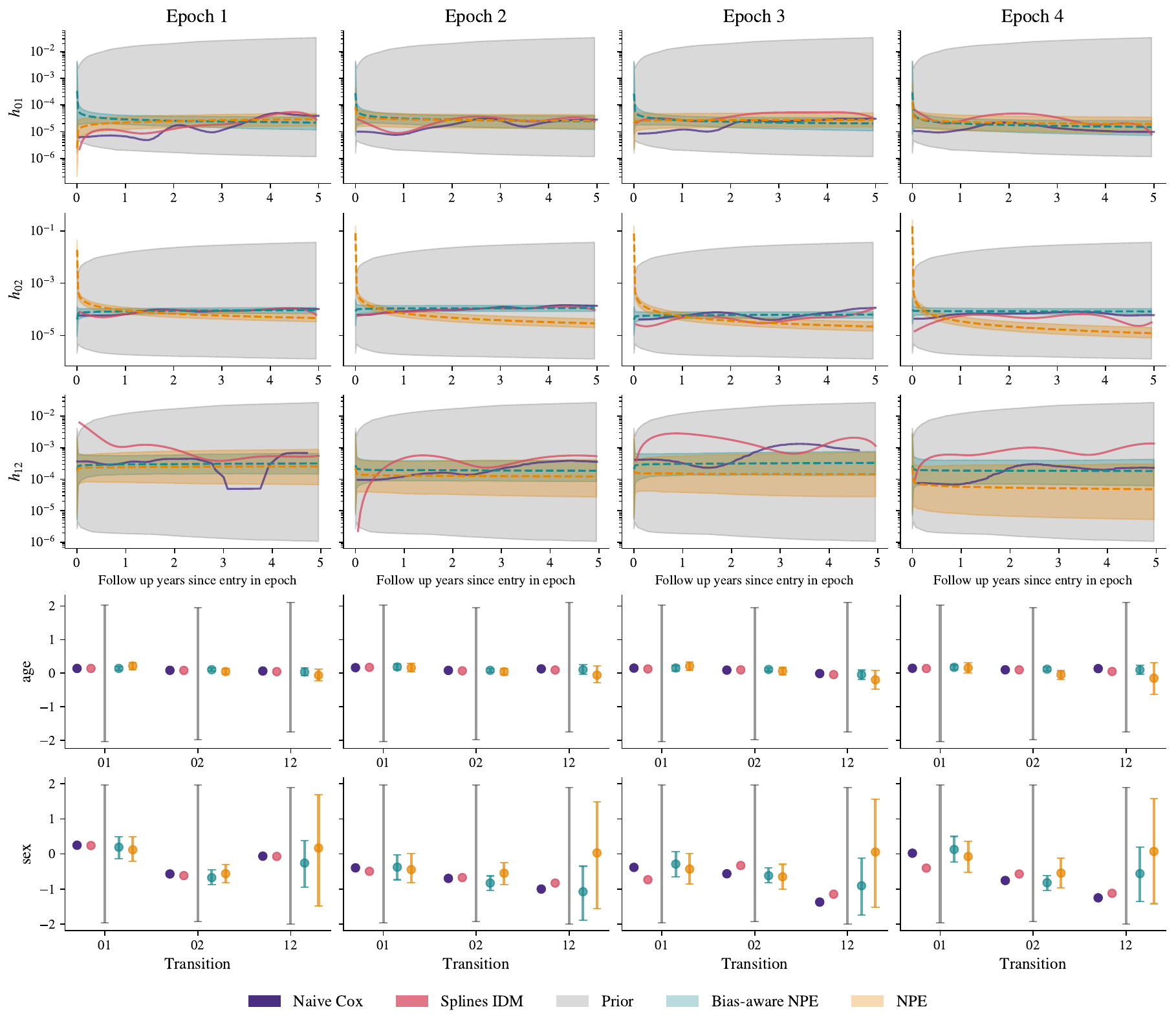}
\caption{
\emph{Additional results for the Framingham Study.}
Unadjusted hazards and covariate effect estimates of naive Cox, penalized likelihood approach, and the posterior for the bias-aware NPE and the NPE trained on full data (median and $95\%$ credible intervals) are compared on the real data. The $95\%$ quantiles of the prior are shown in grey.
}
\label{fig:appendix_cens_vist_recovery}
\end{figure}

\begin{table}[ht]
\centering
\caption{
\emph{Additional results for the Framingham Study.}
Contraction of the posterior for the bias-aware NPE (applied to simulated observed data) and the NPE trained on full data (applied to simulated full data). Median and median absolute deviation over 1000 datasets not used during training are calculated.
}
\label{tab:appendix_cens_visit_contraction}
\begin{tabular}{ccc}
\toprule
Parameter & Bias-aware NPE & NPE \\
\midrule
$a_{01}$ & 0.59 ($\pm$ 0.23) & 0.91 ($\pm$ 0.07) \\
$a_{02}$ & 0.93 ($\pm$ 0.06) & 0.92 ($\pm$ 0.06) \\
$a_{12}$ & 0.33 ($\pm$ 0.28) & 0.32 ($\pm$ 0.30) \\
$\kappa_{01}$ & 0.77 ($\pm$ 0.03) & 0.95 ($\pm$ 0.01) \\
$\kappa_{02}$ & 0.97 ($\pm$ 0.01) & 0.96 ($\pm$ 0.01) \\
$\kappa_{12}$ & 0.79 ($\pm$ 0.03) & 0.73 ($\pm$ 0.03) \\
$\beta_{01}^\text{sex}$ & 0.97 ($\pm$ 0.01) & 0.99 ($\pm$ 0.01) \\
$\beta_{02}^\text{sex}$ & 0.99 ($\pm$ 0.00) & 0.98 ($\pm$ 0.01) \\
$\beta_{12}^\text{sex}$ & 0.89 ($\pm$ 0.07) & 0.22 ($\pm$ 0.04) \\
$\beta_{01}^\text{age}$ & 0.99 ($\pm$ 0.00) & 0.99 ($\pm$ 0.00) \\
$\beta_{02}^\text{age}$ & 0.99 ($\pm$ 0.00) & 0.99 ($\pm$ 0.00) \\
$\beta_{12}^\text{age}$ & 0.99 ($\pm$ 0.01) & 0.99 ($\pm$ 0.01) \\
\bottomrule
\end{tabular}
\end{table}

\begin{figure}[ht!]
\centering
\begin{subfigure}{\linewidth}   
    \centering
    \begin{overpic}[width=\linewidth]{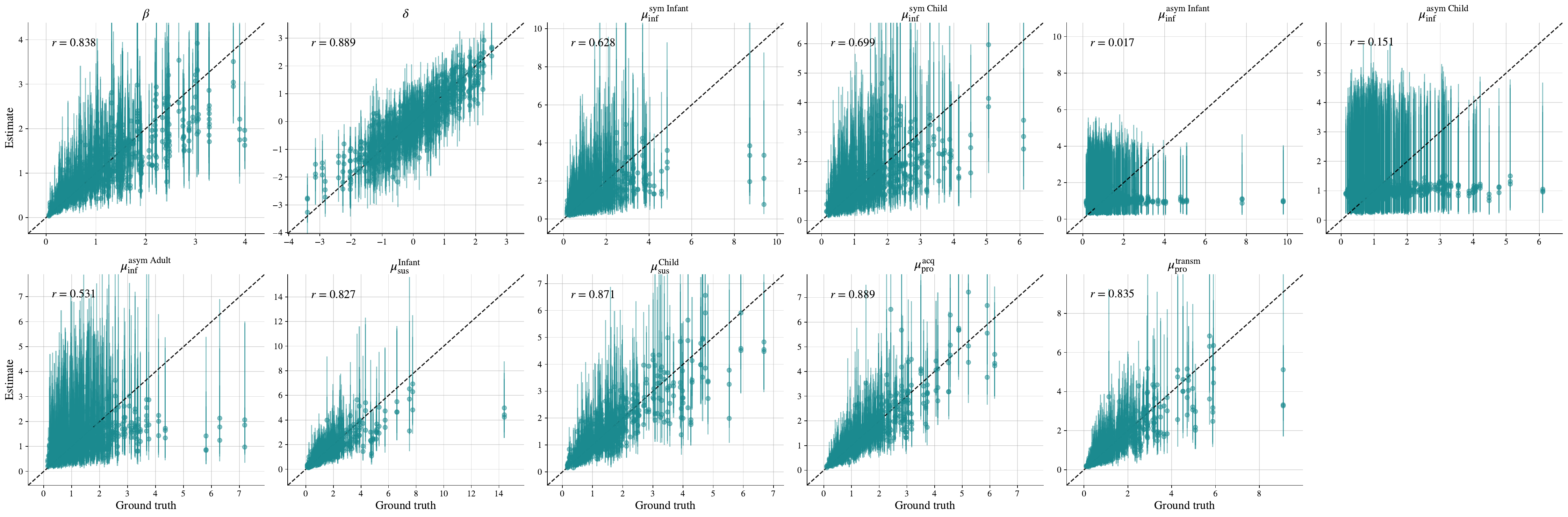}
        \put(-5,145){\textbf{A}}
    \end{overpic}
\end{subfigure}
\begin{subfigure}{\linewidth}   
    \centering
    \begin{overpic}[width=\linewidth]{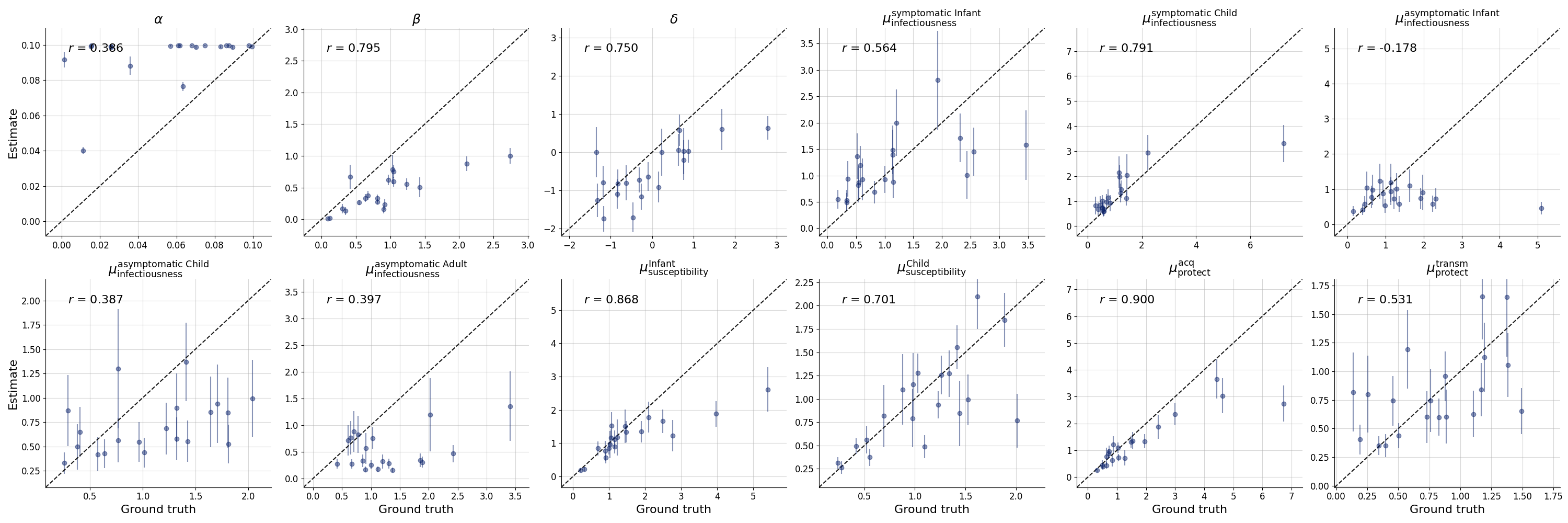}
        \put(-5,145){\textbf{B}}
    \end{overpic}
\end{subfigure}
\caption{
\emph{Additional results for the PedCovid Study.}
(\textbf{A}) Recovery of the parameters on simulated data for all three selection mechanisms using bias-aware NPE. The median and median absolute deviation for each dataset are computed.
(\textbf{B}) Recovery of the parameters on simulated data only for random selection and a unfixed background hazard $\alpha$ using MCMC.
}
\label{fig:appendix_pedcvo_recovery}
\end{figure}

\end{document}